\documentclass{article}
     \PassOptionsToPackage{numbers, compress}{natbib}

\usepackage[final]{neurips_2024}

\usepackage[utf8]{inputenc} %
\usepackage[T1]{fontenc}    %
\usepackage{hyperref}       %
\usepackage{url}            %
\usepackage{booktabs}       %
\usepackage{amsfonts}       %
\usepackage{nicefrac}       %
\usepackage{microtype}      %
\usepackage[table]{xcolor}         %
\usepackage{tikz}
\usepackage{caption}
\usepackage{subcaption}
\usepackage{inconsolata}
\usepackage{graphicx}
\usepackage{amssymb}
\usepackage{bm}
\usepackage{bbm}
\usepackage{amsmath}  %
\usepackage{amsthm}
\usepackage{xcolor, color}
\usepackage{textcomp}
\usepackage{multicol}
\usepackage{multirow}
\usepackage{mathtools}
\usepackage{wrapfig}
\usepackage{mathtools}
\usepackage{tcolorbox}
\usepackage{pgfplots}
\usepackage{enumitem}
\usepackage{filecontents}
\usepackage{pifont}
\usepackage{algorithm}
\usepackage[noend]{algpseudocode}
\usepackage{tikz-layers}
\usepackage{fontawesome}
\usepackage[frozencache,cachedir=.]{minted} %

\algblock[noend]{ParFor}{EndParFor}
\algnewcommand\algorithmicparfor{\textbf{parfor}}
\algnewcommand\algorithmicpardo{\textbf{do}}
\algnewcommand\algorithmicendparfor{\textbf{end\ parfor}}
\algrenewcommand\alglinenumber[1]{}
\algrenewtext{ParFor}[1]{\algorithmicparfor\ #1\ \algorithmicpardo}
\algrenewtext{EndParFor}{\algorithmicendparfor}
\algtext*{EndParFor}
\algnewcommand\algorithmicinput{\textbf{Define}}
\algnewcommand\Def{\item[\algorithmicinput]}

\tcbuselibrary{theorems}
\usetikzlibrary{arrows.meta,positioning,calc,shapes.geometric,decorations.text,patterns,patterns.meta,fit,backgrounds,chains,shadows,math,circuits.ee.IEC,decorations.pathmorphing, decorations.pathreplacing, decorations.shapes}

\pgfdeclarelayer{background}  %
\pgfdeclarelayer{above}       %
\pgfsetlayers{background,main,above}  %

\definecolor{brickred}{HTML}{b92622}
\definecolor{midnightblue}{HTML}{005c7f}
\definecolor{salmon}{HTML}{f1958d}
\definecolor{burntorange}{HTML}{f19249}
\definecolor{junglegreen}{HTML}{4dae9d}
\definecolor{forestgreen}{HTML}{499c5e}
\definecolor{pinegreen}{HTML}{3d8a75}
\definecolor{seagreen}{HTML}{6bc1a2}
\definecolor{limegreen}{HTML}{97c65a}
\definecolor{violet}{HTML}{8f00ff}
\definecolor{pastelviolet}{HTML}{cb99c9}
\definecolor{darkcyan}{HTML}{008B8B}

\newcommand{\suda}{$^1$}
\newcommand{\miteecs}{$^2$}
\newcommand{\ucsc}{$^3$}
\newcommand{\tencent}{$^4$}
\newcommand{\luxi}{$^5$}
\newcommand{\waterloo}{$^6$}
\newcommand{\github}{\raisebox{-0pt}{\faGithub}}

\newcommand{\huggingface}{\raisebox{-2pt}{\includegraphics[scale=0.04]{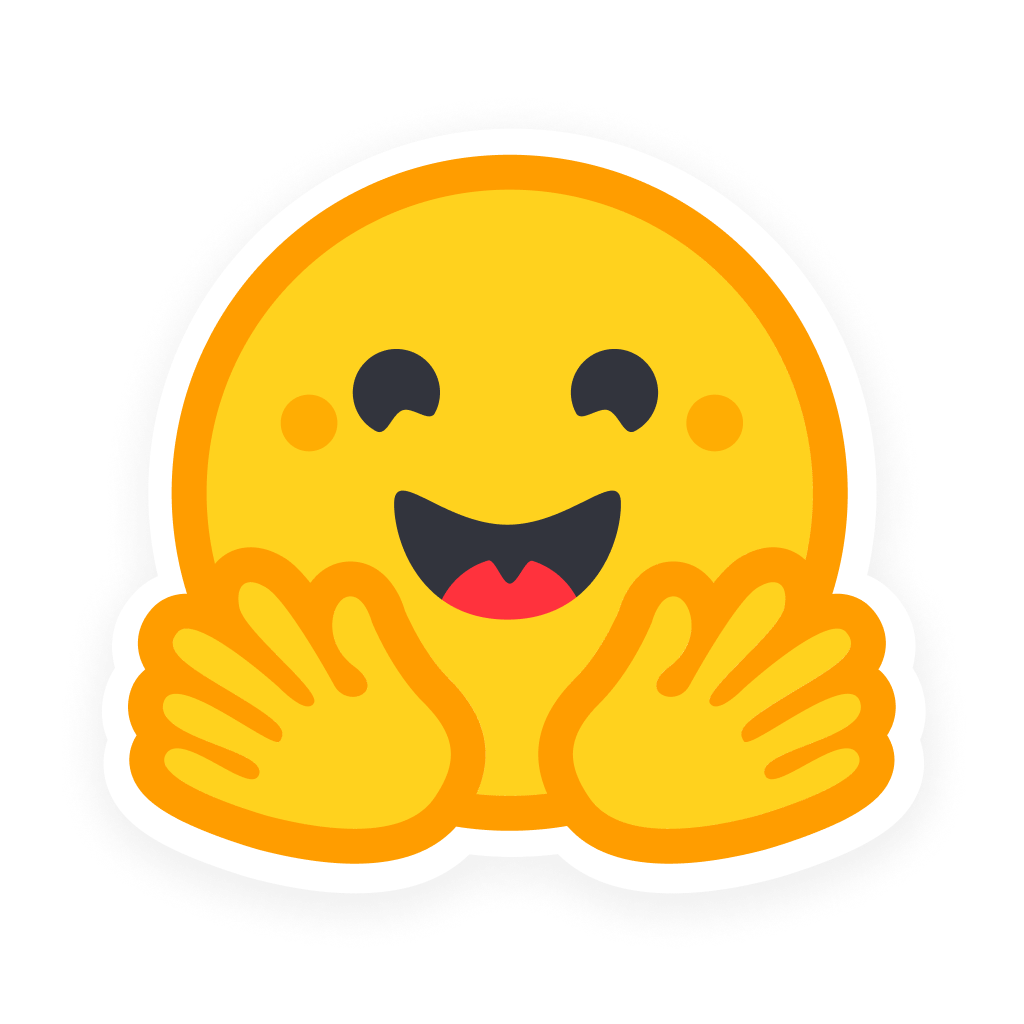}}}

\makeatletter
\renewcommand*{\@fnsymbol}[1]{\ensuremath{\ifcase#1\or * \or \dagger\or \ddagger\or
   \mathsection\or \mathparagraph\or \|\or **\or \dagger\dagger
   \or \ddagger\ddagger \else\@ctrerr\fi}}
\makeatother

\title{Gated Slot Attention for Efficient Linear-Time Sequence Modeling}

\author{%
  Yu Zhang\suda\thanks{Equal contributions. Work was conducted during Yu Zhang's internship at Tencent AI Lab.
}
  \quad
  Songlin Yang\miteecs$^{*}$
  \quad
  Ruijie Zhu\ucsc
  \quad
  Yue Zhang\suda
  \quad
  Leyang Cui\tencent\\
  \textbf{Yiqiao Wang}\luxi
  \quad
  \textbf{Bolun Wang}\luxi
  \quad
  \textbf{Freda Shi}\waterloo
  \quad
  \textbf{Bailin Wang}\miteecs\\
  \quad
  \textbf{Wei Bi}\tencent
  \quad
  \textbf{Peng Zhou}\luxi$^\dagger$
  \quad
  \textbf{Guohong Fu}\suda\thanks{Corresponding authors.}
  \\
\suda School of Computer Science and Technology, Soochow University, China\\
\miteecs Massachusetts Institute of Technology\quad \ucsc University of California, Santa Cruz \\
\tencent Tencent AI Lab\quad \luxi LuxiTech \quad \waterloo University of Waterloo\\
\href{mailto:yzhang.cs@outlook.com}{$\mathtt{yzhang.cs@outlook.com}$} \quad\href{mailto:yangsl66@mit.edu}{$\mathtt{yangsl66@mit.edu}$}\\\\[-8pt]
\github\,\,\url{https://github.com/sustcsonglin/flash-linear-attention}\\
\huggingface\,\,\url{https://huggingface.co/fla-hub}
}

\usepackage{amsmath,amsfonts,bm}

\def\eqref#1{equation~\ref{#1}}

\def\1{\bm{1}}

\def\rmS{{\mathbf{S}}}

\DeclareMathAlphabet{\mathsfit}{\encodingdefault}{\sfdefault}{m}{sl}
\SetMathAlphabet{\mathsfit}{bold}{\encodingdefault}{\sfdefault}{bx}{n}

\newcommand{\R}{\mathbb{R}}

\newcommand{\sigmoid}{\sigma}

\usepackage{todonotes}
\usepackage{marginnote}

\definecolor{tticblue}{RGB}{0, 94, 184}

\begin{document}

\maketitle
\begin{abstract}
    Linear attention Transformers and their gated variants, celebrated for enabling parallel training and efficient recurrent inference, still fall short in recall-intensive tasks compared to traditional Transformers and demand significant resources for training from scratch.
    This paper introduces Gated Slot Attention (GSA), which enhances Attention with Bounded-memory-Control (ABC \cite{peng-etal-2022-abc}) by incorporating a gating mechanism inspired by Gated Linear Attention (GLA \cite{yang-etal-2024-gla}).
    Essentially, GSA comprises a two-layer GLA linked via $\operatorname{softmax}$, utilizing context-aware memory reading and adaptive forgetting to improve memory capacity while maintaining compact recurrent state size.
    This design greatly enhances both training and inference efficiency through GLA's hardware-efficient training algorithm and reduced state size.
    Additionally, retaining the $\operatorname{softmax}$ operation is particularly beneficial in ``finetuning pretrained Transformers to RNNs'' (T2R \cite{kasai-etal-2021-finetuning}) settings, reducing the need for extensive training from scratch.
    Extensive experiments confirm GSA's superior performance in scenarios requiring in-context recall and in T2R settings.
\end{abstract}

\section{Introduction}
Transformers \citep{vaswani-2017-attention} have emerged as the predominant architecture for \emph{most, if not all}, sequence modeling tasks.
Nevertheless, the quadratic complexity of $\operatorname{softmax}$-based standard attention (SA) poses significant challenges for long sequence modeling (e.g., video understanding and biological sequence modeling).
In the context of language modeling, where sequence lengths are moderate, training efficiency is generally not a primary concern.
However, during inference, the Key-Value (KV) cache \cite{irie-2020-how, pope-2022-efficiently} grows linearly with the generation length, resulting in substantial memory burdens and throughput bottlenecks due to high I/O costs.

Linear (kernelized) attention \citep{katharopoulos-2020-transformers} and its gated variants \citep{yang-etal-2024-gla, sun-2023-retnet, qin-2024-transnormerllm, peng-2024-eagle, mamba2, qin-2024-hgrn2} have received interest as promising alternatives to $\operatorname{softmax}$ attention.
These models demonstrate strong performance in language modeling and understanding tasks.
Notably, they can be reframed as RNNs during inference, achieving constant memory complexity and thereby significantly enhancing inference efficiency.

However, two key issues persist with these models:
(i) Performance-wise, recent research indicates that linear recurrent models still struggle with tasks requiring in-context retrieval or learning \citep{arora-2023-zoology, akyurek-etal-2024-inctx, DBLP:conf/icml/JelassiBKM24, Grazzi2024IsMC}, and there is a \emph{fundamental} recall-memory trade-off \citep{arora-2024-simple, Wen2024RNNsAN} where all inference-time-constant-memory models face inherent limitations.
(ii) In terms of training efficiency, while linear attention supports hardware-efficient chunkwise training \citep{yang-etal-2024-gla} as implemented in FlashLinearAttention (FLA \citep{yang-2024-fla}), training from scratch on trillions of tokens remains prohibitively expensive.
A paradigm, ``\emph{finetuning pretrained Transformers to RNNs}'' (short for T2R \citep{kasai-etal-2021-finetuning}), has recently gained great attention \citep{zhang-2024-hedgehog, chen-2024-dijiang, mercat-2024-linearizing, choi-2024-cross, bick-2024-transformertossms, wang-2024-mamballama}.
This approach circumvents the high cost of training from scratch by requiring only a few billion tokens for finetuning—about 1–3$\%$ of the total cost.
However, linear attention uses a different kernel method from $\operatorname{softmax}$, leading to performance discrepancies when finetuning pretrained $\operatorname{softmax}$ attention models to linear attention \citep{zhang-2024-hedgehog}.

To address these issues, we revisit the Attention with Bounded-Memory Control (ABC) model \citep{peng-etal-2022-abc}, which retains the $\operatorname{softmax}$ operation, thereby reducing training-finetuning discrepancies between standard and linear attention, making it ideal for T2R settings.
Additionally, ABC enables more effective state utilization, requiring less state size to achieve similar performance, as observed in \citet{peng-etal-2022-abc}.
This results in more efficient inference and potentially expands the Pareto frontier of the recall-memory tradeoff \citep{arora-2024-simple}.
However, ABC has not gained significant attention due to its mediocre language modeling performance and slow training speed.

In this work, we first reformulate ABC as two-pass linear attention linked via $\operatorname{softmax}$, allowing us to leverage the hardware-efficient chunkwise implementation from FLA \citep{yang-2024-fla} for more efficient training.
We then identify several limitations of ABC and propose a new model, dubbed Gated Slot Attention (\textsc{GSA}), which is essentially a gated version of ABC, following the recent trend of enhancing linear attention with gating mechanisms \citep{yang-etal-2024-gla, qin-2024-hgrn2, peng-2024-eagle, mamba2, beck-2024-xlstm, peng-2021-rfa,mao-2022-fine, pramanik-2023-recurrentlineartransformers}.

Our extensive evaluation shows that \textsc{GSA} not only matches performance in language modeling and understanding tasks but also significantly outperforms other linear models in \emph{in-context recall-intensive} tasks \citep{arora-2024-simple, arora-2024-jrt}, without requiring a large state size like RetNet \citep{sun-2023-retnet} or GLA \citep{yang-etal-2024-gla}.
In the T2R finetuning setting, we found that finetuning Mistral-7B \citep{jiang-2023-mistral} to \textsc{GSA} surpasses large recurrent language models (e.g., RWKV6-7B, Mamba-7B) and also outperforms finetuning Mistral-7B to other linear models (e.g., RetNet, GLA) and other T2R methods like SUPRA \cite{mercat-2024-linearizing}, verifying the importance of retaining the $\operatorname{softmax}$ operator.
Finally, we remark that \textsc{GSA} achieves similar training speeds to GLA while offering an inference speedup due to its smaller state size.

\section{Background and Preliminary}

\subsection{Transformers as Unbounded Key-Value Memories}
Given $\mathbf{X} = \left[\boldsymbol{x}_1,\dots, \boldsymbol{x}_T\right]^\top \in \mathbb{R}^{T \times d}$, where $T$ is the sequence length and $\boldsymbol{x}_i\in \mathbb{R}^d$ is the $i$-th input vector with $d$ dimensions, SA with causal masking computes the output matrix:
\begin{equation}
  \label{eq:standard-attention}
  \mathbf{O} = f((\mathbf{Q} \mathbf{K}^\top)\odot\mathbf{M})\mathbf{V},
\end{equation}
where $\mathbf{Q, K, V}\in \mathbb{R}^{T\times d}$ are linear mappings of the input $\mathbf{X}$ via learnable weights $\mathbf{W}_q,\mathbf{W}_k,\mathbf{W}_v \in \mathbb{R}^{d\times d}$,
$\textbf{M} = \left\{M_{ij} = 1 \text{ if } i \geq j \text{ o.w. }  -\infty\right\}$ is the causal mask to prevent future information leakage, $\odot$ denotes element-wise production,
and $f(\cdot)$ is
$\operatorname{softmax}\left(\cdot\right)$.

Generally, $\mathbf{K},\mathbf{V}$ can be viewed as neural \emph{key-value memories} $\widetilde{\mathbf{K}}_t,\widetilde{\mathbf{V}}_t\in \mathbb{R}^{m\times d}$, respectively \citep{sukhbaatar-2015-memory,geva-etal-2021-transformer}, where $m$ is the number of memory slots.
At step $t$, the query $\boldsymbol{q}_t=\mathbf{W}_q \boldsymbol{x}_t \in \mathbb{R}^{d}$ first attends to the key memories $\widetilde{\mathbf{K}}_t$ to retrieve relevant information, which is then summarized into $\boldsymbol{o}_t$ by computing a weighted sum of the value memories $\widetilde{\mathbf{V}}_t$ \citep{zhang-cai-2022-linearizing}, where the weights are the normalized attention scores:
\begin{equation} \label{eq:memory}
  \boldsymbol{o}_t = {\widetilde{\mathbf{V}}_t}^\top f({\widetilde{\mathbf{K}}_t} \boldsymbol{q}_t).
\end{equation}
From this perspective, Transformers are equipped with an unbounded number of memory slots, which grow linearly with respect to the sequence length \citep{oren-2024-transformers} (i.e., $m = t$ for step $t$)---a new key $\boldsymbol{k}_t=\mathbf{W}_k \boldsymbol{x}_t \in \mathbb{R}^{d}$ is assigned with a unique memory slot upon its introduction.
This leads to a simple memory updating rule: $\widetilde{\mathbf{K}}_t = \widetilde{\mathbf{K}}_{t-1} \cup \{\boldsymbol{k}_t\}$.
The value memories $\widetilde{\mathbf{V}}_t$ are updated in a similar way.
This mechanism, however, comes at the cost of quadratic time complexity in terms of the sequence length for training and
$O(T d)$ time/memory complexity for inference \citep{pope-2022-efficiently}, posing challenges for large-scale models.

\subsection{ABC \citep{peng-etal-2022-abc}: Linearizing Attention with Bounded Memory Control}
\label{sec:abc}

From a key-value memory perspective, the training and inference complexity of self-attention (SA) can be reduced by fixing the number of memory slots to a constant size \(m \ll T\) \citep{graves-2014-neural, ma-2021-luna, peng-etal-2022-abc}.
One straightforward way to achieve this is by employing a \emph{first-in-first-out} memory management strategy, commonly known as sliding window attention (SWA).
However, SWA is inefficient because it discards all information outside the window, leading to poor performance in balancing the recall-memory tradeoff \cite{arora-2024-simple}.
To achieve acceptable performance, SWA often requires a large window size (e.g., 4,096 tokens in Mistral \citep{jiang-2023-mistral}), which diminishes its advantage over to global attention.

When the number of tokens in a sequence exceeds the number of memory slots, it becomes necessary to store information from multiple tokens in a single slot.
To address this challenge, \citet{peng-etal-2022-abc} propose the Attention-with-Bounded-memory-Control (ABC) mechanism, which allows multiple tokens to be written into a single slot:
\begin{equation}
  \label{eq:abc_recurrent}
  \widetilde{\mathbf{K}}_{t}=\widetilde{\mathbf{K}}_{t-1}+\boldsymbol{\phi}_{t} \otimes \boldsymbol{k}_{t} \in \mathbb{R}^{m\times d}, \quad  \widetilde{\mathbf{V}}_{t}=\widetilde{\mathbf{V}}_{t-1}+\boldsymbol{\phi}_{t} \otimes \boldsymbol{v}_{t} \in \mathbb{R}^{m \times d}, \quad \mathbf{o}_t =\widetilde{\mathbf{V}}^T f( \widetilde{\mathbf{K}}_t^T \mathbf{q}_t ) \in \mathbb{R}^d
\end{equation}
where
\begin{equation}
  \boldsymbol{\alpha}_i = \exp \left(\mathbf{W}_\phi \mathbf{x}_i\right) \in \mathbb{R}^{m}, \quad \boldsymbol{\phi}_i = \frac{\boldsymbol{\alpha}_i}{\sum_{j=1}^i \boldsymbol{\alpha}_j} \in (0,1)^{m}
  \label{eq:abc_phi}
\end{equation}
Here, \((\boldsymbol{\phi}_i)_j\) represents the writing intensity of the \(i\)th token to the \(j\)th slot, obtained using a cumulative $\operatorname{softmax}$ function (cf. \citep[][footnote 5]{peng-etal-2022-abc}),  which can be computed with a prefix sum.

\paragraph{ABC as two-pass linear attention.}

The outer-product-based additive memory update rule in Eq.~\ref{eq:abc_recurrent} bears a resemblance to linear attention \citep{katharopoulos-2020-transformers}, which involves the following recurrence\footnote{For simplicity, we omit the normalization term, which has been shown to be unnecessary \citep{schlag-2021-deltanet, qin-etal-2022-devil,mao-2022-fine,sun-2023-retnet,yang-etal-2024-gla}.}:
\begin{align}
  \mathbf{S}_t = \mathbf{S}_{t-1} + \boldsymbol{k}_t \otimes \boldsymbol{v}_t \in \mathbb{R}^{d\times d}, \quad\quad\boldsymbol{o}_t = \mathbf{S}^T_t \boldsymbol{q}_t \in \mathbb{R}^{d}
  \label{eq:la}
\end{align}
We denote this linear attention operator that computes $\boldsymbol{o}_i$ from $\boldsymbol{q}_i, \boldsymbol{k}_i$ and $\boldsymbol{v}_i$ (Eq.~\ref{eq:la}) by $\{\boldsymbol{o}_i\}_{i=1}^T = \operatorname{LA}(\{\boldsymbol{q}_i, \boldsymbol{k}_i, \boldsymbol{v}_i\}_{i=1}^{T})$.
We show that the ABC operations can be written as
\begin{align*}
  \{\boldsymbol{o}^{\prime}_i\}_{i=1}^T & = \operatorname{LA}(\{\boldsymbol{q}_i,  \boldsymbol{k}_i, \boldsymbol{\phi}_i\}_{i=1}^T),                                 \\
  \{\boldsymbol{o}_i\}_{i=1}^T          & = \operatorname{LA}(\{\operatorname{softmax}(\boldsymbol{o}^{\prime}_i), \boldsymbol{\phi}_i, \boldsymbol{v}_i\}_{i=1}^T),
\end{align*}
where $\boldsymbol{o}^{\prime}_i \in \mathbb{R}^m, \boldsymbol{o}_i \in \mathbb{R}^d$.
Therefore, ABC can enjoy hardware-efficient linear-time chunkwise training \citep{yang-etal-2024-gla}, as implemented in the FLA library \citep{yang-2024-fla}.

\paragraph{Remarks on state size.}

\citet{peng-etal-2022-abc} empirically demonstrated that ABC requires a smaller state size to achieve comparable performance to other linear attention models, resulting in improved inference efficiency.
We offer the following intuitive explanation: the new query $\boldsymbol{o}^{\prime}$ aggregates the entire history through the initial pass of linear attention, making it more context-aware and better at locating desired items for retrieval.
The subsequent $\operatorname{softmax}$ operator helps mitigate the attention dilution issue \citep{qin-etal-2022-devil}.
From the perspective of Hopfield networks, softmax can exponentially increase the memory size \citep{krotov_large_2021}.
Together, these factors suggest that ABC may possess an implicit large memory capacity, even with a small actual recurrent state size.

\subsection{GLA \citep{yang-etal-2024-gla}: Linear Attention with  Gating Mechanism}
Linear attentions underperform $\operatorname{softmax}$-attention Transformers in language modeling by a notable margin.
RetNet \citep{sun-2023-retnet} and TransnormerLLM \citep{qin-2024-transnormerllm} incorporate a \emph{data-independent} exponential decay factor for memory update as
\begin{align*}
  {\mathbf{S}}_{t} & = \gamma {\mathbf{S}}_{t-1}+ \boldsymbol{k}_{t} \otimes \boldsymbol{v}_{t} \in \mathbb{R}^{d\times d},
\end{align*}
where $\gamma \in (0,1)$ is a scalar data-independent decaying factor; that is, the decay rate is fixed across time steps and hidden channels (under the same head), disrespect to the input tokens.
RetNet has shown better language modeling performance compared to vanilla linear attentions thanks to the decaying mechanism.

However, research in recurrent neural networks (RNNs) has shown that \emph{data-dependent} decay (or forget gates) is crucial for selectively retaining and forgetting information \citep{gers-1999-forget,graves-2014-generating}, thus better leveraging the fixed recurrent hidden state.
This selective mechanism has been revisited in recent state-space models \citep{gu-2023-mamba, mamba2}.
Inspired by LSTMs, Gated Linear Attention (GLA) \citep{mao-2022-fine, yang-etal-2024-gla} introduces data-dependent decay parameters $\mathbf{G}_t \in (0,1)^{d\times d}$ to gate the hidden state as follows,
\begin{align*}
  \rmS_t & = \mathbf{G}_t \odot  \rmS_{t-1} + \boldsymbol{k}_t \otimes \boldsymbol{v}_t \in \mathbb{R}^{d\times d}, \quad\boldsymbol{o}_t  = \rmS_t^T \boldsymbol{q}_t  \in \mathbb{R}^d.
\end{align*}
\cite{yang-etal-2024-gla} show that if gates are parameterized in an outer product form $\mathbf{G}_t = \boldsymbol{\alpha}_t \otimes \boldsymbol{\beta}_i$, and $\boldsymbol{\alpha}_t, \boldsymbol{\beta}_t \in [0, 1]^{d}$ depend solely on input $\boldsymbol{x}_t$,
such recurrence can be rewritten as matrix multiplication, allowing for hardware-efficient training with a chunkwise parallel form.
In what follows, we will use the following notation
$\operatorname{GLA}(\{\boldsymbol{q}_i,\boldsymbol{k}_i,\boldsymbol{v}_i,\boldsymbol{\alpha}_i,\boldsymbol{\beta}_i\}_{i=1}^T) = \{\boldsymbol{o}_i\}_{i=1}^T$ to denote this computation.
It is common to set $\boldsymbol{\beta}_i = \mathbf{1}$ as in \cite{yang-etal-2024-gla, qin-2024-hgrn2, peng-2024-eagle}, which is also often written in the following equivalent form:
\begin{align*}
  \mathbf{S}_t = \operatorname{Diag}(\boldsymbol{\alpha}_t) \mathbf{S}_{t-1} + \boldsymbol{k}_t \otimes \boldsymbol{v}_t.
\end{align*}
Here $\boldsymbol{k}_t$ can be viewed as the input gate, and $\boldsymbol{\alpha}_t$ can be viewed as the forget gate.
In gated RNN literature, it is common to couple these two gates via $\boldsymbol{k}_t = 1-\boldsymbol{\alpha}_t$ \citep{cho-etal-2014-gru, zhou-2016-mgu, qin-2023-hgrn}.
In particular, \citet{qin-2024-hgrn2} proposed HGRN2, which uses this strategy as an improved parameterization of GLA, showing better performance in language modeling.

\section{Method}

\subsection{Motivation: Issues with ABC}

We identify two primary limitations in ABC's memory update rule.
Firstly, it lacks a forgetting mechanism, resulting in indefinite retention of items once written into memory slots.
This prevents efficient memory reuse by impeding the prompt clearance of slots for new information.

Secondly, the rule introduces an unwarranted inductive bias favoring tokens at the sentence's beginning.
This contradicts the recency bias in natural language, where more recent information is often more relevant.
Prioritizing initial tokens over the recent ones conflicts with this inherent tendency in natural language processing.

Specifically, for the first token, the writing strength to all slots is maximized (i.e., \(\phi_1 = \mathbf{1} \in \mathbb{R}^m\)), causing every memory slot to retain a copy of the first token's representation.
The absence of a forgetting mechanism exacerbates this issue.
For subsequent tokens, the writing strength diminishes due to the influence of earlier tokens, as a result of the cumulative $\operatorname{softmax}$ in Eq.~\ref{eq:abc_phi}.
This makes it challenging for the model to retain later tokens without learning a significantly large \(\alpha_i\), potentially leading to instability in long-context settings, as observed by \citet{zhang-2022-cab}.

\subsection{Gated Slot Attention (\textsc{GSA}): ABC with gating mechanism}
To address these limitations, we propose Gated Slot Attention (\textsc{GSA}), which incorporates a gating mechanism to simultaneously resolve both issues by: (i) enabling the forgetting of historical information, and
(ii) introducing a recency inductive bias, as detailed below.

For each memory slot, the update rule is a simple gated RNN with a scalar data-dependent gating value $\alpha_i \in [0, 1]$,
\begin{align*}
  (\widetilde{\mathbf{K}}_t)_i = \alpha_i (\widetilde{\mathbf{K}}_{t-1})_i + (1-\alpha_i) \boldsymbol{k}_t \in \mathbb{R}^d, \qquad (\widetilde{\mathbf{V}}_t)_i = \alpha_i (\widetilde{\mathbf{V}}_{t-1})_i + (1-\alpha_i) \boldsymbol{v}_t \in \mathbb{R}^d
\end{align*}
and these can be written in matrix form,  which is reminiscent of HGRN2 \cite{qin-2024-hgrn2}.
\begin{equation}
  \label{eq:gsa}
  \begin{aligned}
    \widetilde{\mathbf{K}}_{t} & = \operatorname{Diag}(\boldsymbol{\alpha}_{t}) \cdot   \widetilde{\mathbf{K}}_{t-1}+ (1-\boldsymbol{\alpha}_{t}) \otimes \boldsymbol{k}_{t} \in \mathbb{R}^{m\times d}  \\
    \widetilde{\mathbf{V}}_{t} & = \operatorname{Diag}(\boldsymbol{\alpha}_{t}) \cdot   \widetilde{\mathbf{V}}_{t-1}+ (1-\boldsymbol{\alpha}_{t}) \otimes \boldsymbol{v}_{t} \in \mathbb{R}^{m \times d}
    \\
    \mathbf{o}_t               & =\widetilde{\mathbf{V}}^T \operatorname{softmax}( \widetilde{\mathbf{K}}_t^T \mathbf{q}_t ) \in \mathbb{R}^d
  \end{aligned}
\end{equation}

\paragraph{GSA as two-pass GLA.} It is straightforward to see that we can write GSA as a two-pass GLA as shown below:
\begin{equation}\label{eq:gsa-2pass}
  \begin{aligned}
    \{\boldsymbol{o}^{\prime}_t\}_{t=1}^T & = \operatorname{GLA}\left(\{\boldsymbol{q}_t,   \boldsymbol{k}_t, 1-\boldsymbol{\alpha}_{t}, \boldsymbol{\alpha}_t, \mathbf{1}\}^T_{t=1}\right)                                   \\
    \{\boldsymbol{o}_t\}_{t=1}^T          & = \operatorname{GLA}\left(\{\operatorname{softmax}(\boldsymbol{o}^{\prime}_t), 1-\boldsymbol{\alpha}_{t}, \boldsymbol{v}_t, \mathbf{1}, \boldsymbol{\alpha}_t \}_{t=1}^{T}\right)
  \end{aligned}
\end{equation}
Therefore, we can adapt GLA’s hardware-efficient chunkwise training algorithm for GSA training, as shown in \S~\ref{appendix:la} and \S~\ref{appendix:alg}.
We illustrate the recurrent representation of \textsc{GSA} in Figure~\ref{fig:gsa-recurrent}.

\subsection{Neural Architecture}
\begin{figure}[h!]
  \begin{minipage}{0.4\textwidth}
    \centering
    \vspace{-7mm}
    \definecolor{kcell_color}{RGB}{203,231,207}
    \definecolor{vcell_color}{RGB}{220,223,240}
    \definecolor{gray_bbox_color}{RGB}{243,243,244}
    \definecolor{operator_color}{RGB}{255,220,220}
    \resizebox{.95\textwidth}{!}{
      \begin{tikzpicture}[
          font=\small,
          kcell/.style={
              draw=black,
              very thick,
              fill=kcell_color!30,
              rounded corners=10pt
            },
          vcell/.style={
              draw=black,
              very thick,
              fill=vcell_color!40,
              rounded corners=10pt
            },
          inputbox/.style={
              very thick,
              rounded corners=3pt
            },
          function/.style={
              draw,
              very thick,
              fill=operator_color!80,
              rounded corners=3pt,
              inner sep=0pt,
              outer sep=0pt,
              minimum size=11pt,
              minimum width=30pt,
            },
          control/.style={
              draw,
            },
          silu/.style={
              draw=black,
              very thick,
              fill=silu_color,
              minimum width=1.1cm,
              rounded corners=3pt
            },
          oplus/.style={
              draw=black,
              very thick,
              circle,
              fill=operator_color!80,
              minimum size=11pt,
              inner sep=0pt,
              outer sep=0pt,
              path picture={
                  \node at (path picture bounding box.center) {$\boldsymbol{+}$};
                }
            },
          otimes/.style={
              draw=black,
              very thick,
              circle,
              fill=operator_color!80,
              minimum size=11pt,
              inner sep=0pt,
              outer sep=0pt,
              path picture={
                  \node at (path picture bounding box.center) {$\boldsymbol{\times}$};
                }
            },
          odot/.style={
              draw=black,
              very thick,
              circle,
              fill=operator_color!80,
              minimum size=11pt,
              inner sep=0pt,
              outer sep=0pt,
              path picture={
                  \node at (path picture bounding box.center) {$\boldsymbol{\times}$};
                }
            },
          oneminus/.style={
              draw=black,
              very thick,
              circle,
              fill=operator_color!80,
              minimum size=11pt,
              inner sep=0pt,
              outer sep=0pt,
              path picture={
                  \node at (path picture bounding box.center) {\tiny$\operatorname{1-}$};
                }
            },
          sigmoid/.style={
              draw=black,
              line width=1pt,
              circle,
              fill=operator_color!80,
              minimum size=11pt,
              inner sep=0pt,
              outer sep=0pt,
              path picture={
                  \node at (path picture bounding box.center) {$\mathtt{\sigma}$};
                }
            },
          swish/.style={
              draw=black,
              thick,
              line width=1pt,
              circle,
              fill=operator_color!80,
              minimum size=11pt,
              inner sep=0pt,
              outer sep=0pt,
              path picture={
                  \draw[domain=-1.5:0, samples=50, variable=\x, blue, thick]
                  plot ({\x}, {0});
                  \draw[domain=0:1.5, samples=50, variable=\x, blue, thick]
                  plot ({\x}, {\x});
                }
            },
          forgetgate/.style={
              circuit ee IEC,
              very thick,
              small circuit symbols,
              set make contact graphic= var make contact IEC graphic
            },
          mylabel/.style={
              font=\scriptsize
            },
          link1/.style={
              rounded corners=5pt,
              very thick,
            },
          inputlink/.style={
              decorate,decoration={zigzag,segment length=4pt,amplitude=1.5pt,pre length=3pt,post length=3pt},
            },
          circuit ee IEC,
          very thick,
          small circuit symbols,
          set make contact graphic= var make contact IEC graphic
        ]

        \node [kcell, minimum height=72pt, minimum width=150pt] (cellk) {};
        \node [inputbox, anchor=south, minimum height=30pt, minimum width=150pt] at ($(cellk.north)$) (input) {};
        \node [vcell, anchor=south, minimum height=72pt, minimum width=150pt] at ($(input.north)$) (cellv) {};
        \node [inputbox, anchor=north, minimum height=12pt, minimum width=150pt] at ($(cellk.south)$) (input0) {};
        \node[anchor=east] (k0) at ($(cellk.south west)!0.2!(cellk.north west)+(-0.2,0)$) {$\widetilde{\mathbf{K}}_{t-1}$};
        \node[anchor=west] (k1) at ($(cellk.south east)!0.2!(cellk.north east)+(0.4,0)$) {$\widetilde{\mathbf{K}}_{t}$};
        \node[anchor=east] (v0) at ($(cellv.south west)!0.8!(cellv.north west)+(-0.2,0)$) {$\widetilde{\mathbf{V}}_{t-1}$};
        \node[anchor=west] (v1) at ($(cellv.south east)!0.8!(cellv.north east)+(0.4,0)$) {$\widetilde{\mathbf{V}}_{t}$};
        \node[anchor=east] (i0) at ($(input.south west)!0.5!(input.north west)+(-0.3,0)$) {};

        \node[sigmoid] at ($(input.west)+(0.4,0)$) (fgate) {};
        \node[] (phi) at ($(fgate)+(0.7,0)$) {$\boldsymbol{\alpha}_t$};
        \node[oneminus] (oneminus) at ($(phi)+(0.9,0)$) {};

        \node[] (x) at ($(input)+(0,0)$) {};
        \node[] (xq) at ($(input)+(0.8,0)$) {};
        \node[] (k) at ($(cellk.north)!0.2!(cellk.south)$) {$\boldsymbol{k}_t$};
        \node[] (v) at ($(cellv.south)!0.2!(cellv.north)$) {$\boldsymbol{v}_t$};
        \node[] (q) at ($(k)+(0.8,0)$) {$\boldsymbol{q}_t$};
        \draw [inputlink] (i0) -- (fgate);
        \draw [link1, control] (fgate) -- (phi) -- (oneminus);
        \draw [inputlink] (x) -- (v);
        \draw [inputlink] (x) -- (k);
        \draw [inputlink] (xq) -- (q);

        \node[oplus] (plus1) at ($(k0-|k)$) {};
        \node[oplus] (plus2) at ($(v0-|v)$) {};
        \node[otimes] (otimes1) at ($(cellk.south east-|plus1)!0.5!(cellk.north east-|plus1)$) {};
        \node[otimes] (otimes2) at ($(cellv.south east-|plus2)!0.5!(cellv.north east-|plus2)$) {};

        \node[] (dot3) at ($(k0-|oneminus)$) {};
        \node[] (dot4) at ($(k0-|q)$) {};
        \node[odot] (dot5) at ($(otimes1-|q)+(0.8,0)$) {};
        \node[function] (softmax) at ($(input-|q)+(0.8,0)$) {\tiny$\operatorname{softmax}$};
        \node[] (dot6) at ($(k0-|softmax)$) {};
        \node[odot] (dot7) at ($(dot6|-otimes2)$) {};
        \node[] (output) at ($(dot7-|v1)$) {$\boldsymbol{o}_{t}$};

        \draw [link1] (k0) -- ($(k0-|phi)+(0.4,0)$);
        \draw ($(phi|-k0)+(0.4,0)$) to [make contact] ($(k0-|oneminus)+(-0,0)$);
        \draw ($(v0-|phi)+(0.4,0)$) to [make contact] ($(v0-|oneminus)+(-0,0)$);
        \draw [link1] (v0) -- ($(v0-|phi)+(0.4,0)$);
        \draw [-{Latex[length=6pt,width=5pt]}, link1, control] (phi) |- ($(k0-|phi)+(0.4,0)$);
        \draw [-{Latex[length=6pt,width=5pt]}, link1, control] (phi) |- ($(v0-|phi)+(0.4,0)$);
        \draw [-{Latex[length=6pt,width=5pt]}, link1] ($(k0-|oneminus)$) -- (plus1);
        \draw [-{Latex[length=6pt,width=5pt]}, link1] (plus2) -- (v1);
        \draw [-{Latex[length=6pt,width=5pt]}, link1] ($(v0-|oneminus)$) -- (plus2);

        \draw [-{Latex[length=6pt,width=5pt]}, link1, control] (oneminus) |- (otimes1);
        \draw [-{Latex[length=6pt,width=5pt]}, link1, control] (oneminus) |- (otimes2);
        \draw [-{Latex[length=6pt,width=5pt]}, link1] (k) -- (otimes1);

        \draw [-{Latex[length=6pt,width=5pt]}, link1] (otimes1) -- (plus1);
        \draw [-{Latex[length=6pt,width=5pt]}, link1] (v) -- (otimes2);
        \draw [-{Latex[length=6pt,width=5pt]}, link1] (otimes2) -- (plus2);
        \draw [-{Latex[length=6pt,width=5pt]}, link1] (plus1) -| (dot5);
        \draw [-{Latex[length=6pt,width=5pt]}, link1] (q) |- (dot5);
        \draw [-{Latex[length=6pt,width=5pt]}, link1] (dot5) -- (softmax);
        \draw [-{Latex[length=6pt,width=5pt]}, link1] (softmax) -- (dot7);
        \draw [-{Latex[length=6pt,width=5pt]}, link1] (plus2) -| (dot7);
        \draw [-{Latex[length=6pt,width=5pt]}, link1] (dot7) -- (output);

        \draw [-{Latex[length=6pt,width=5pt]}, link1] (plus1) -- (k1);

      \end{tikzpicture}
    }
    \caption{\small
      The recurrent representation of GSA.
      \protect\tikz[baseline=-0.5ex] \protect\draw[decorate,decoration={zigzag,segment length=4pt,amplitude=1.5pt,pre length=3pt,post length=3pt},very thick] (0,0) -- (0.5,0); means taking $\boldsymbol{x}_t$ as input.
    }
    \label{fig:gsa-recurrent}
    \vspace{-5\normalbaselineskip}
  \end{minipage}
  \begin{minipage}{0.59\textwidth}
    \centering
    \definecolor{fgate_color}{RGB}{252,224,225}
    \definecolor{abc_color}{RGB}{252,226,187}
    \definecolor{add_norm_color}{RGB}{242,243,193}
    \definecolor{glu_color}{RGB}{194,232,247}
    \definecolor{silu_color}{RGB}{203,231,207}
    \definecolor{linear_color}{RGB}{220,223,240}
    \definecolor{gray_bbox_color}{RGB}{243,243,244}
    \definecolor{oproj_color}{RGB}{203,231,207}
    \definecolor{operator_color}{RGB}{252,224,225}
    \resizebox{\textwidth}{!}{
      \tikzset{
        model/.style={
            draw=black,
            very thick,
            fill=gray_bbox_color,
            minimum width=4cm,
            minimum height=5.4cm,
            rounded corners=10pt
          },
        gsa/.style={
            draw=black,
            very thick,
            fill=gray_bbox_color,
            minimum width=5.5cm,
            minimum height=5.4cm,
            rounded corners=10pt
          },
        tokenmixer/.style={
            draw=black,
            very thick,
            fill=abc_color!80,
            minimum width=2.5cm,
            minimum height=0.7cm,
            rounded corners=3pt
          },
        glu/.style={
            draw=black,
            very thick,
            fill=glu_color!80,
            minimum width=2.5cm,
            minimum height=0.7cm,
            rounded corners=3pt
          },
        norm/.style={
            draw=black,
            very thick,
            fill=add_norm_color!80,
            minimum width=2.5cm,
            rounded corners=3pt,
            align=center,
          },
        linear/.style={
            draw=black,
            very thick,
            fill=oproj_color!80,
            minimum width=2.5cm,
            rounded corners=3pt
          },
        stacked/.style={
            draw=black,
            very thick,
            fill=linear_color!80,
            minimum width=2.5cm,
            rounded corners=3pt,
            rectangle,
          },
        fgate/.style={
            draw=black,
            very thick,
            fill=fgate_color!80,
            minimum width=1.1cm,
            rounded corners=3pt
          },
        oproj/.style={
            draw=black,
            very thick,
            fill=oproj_color!80,
            minimum width=2.5cm,
            rounded corners=3pt
          },
        silu/.style={
            draw=black,
            very thick,
            fill=silu_color,
            minimum width=1.1cm,
            rounded corners=3pt
          },
        layerlink/.style={
            -latex,
            very thick,
          },
        modulelink/.style={
            -latex,
            very thick,
            densely dashed,
            shorten >=1pt,
            shorten <=1pt,
            rounded corners=3pt
          },
        normlink/.style={
            very thick,
          },
        residual/.style={
            very thick,
            rounded corners=5pt
          },
        qf/.style={
            -latex,
            very thick,
            rounded corners=5pt
          },
        oplus/.style={
            draw=black,
            line width=1pt,
            circle,
            minimum size=8pt,
            inner sep=0pt,
            outer sep=0pt,
            path picture={
                \draw (path picture bounding box.center) -- ++(0.3cm,0)
                (path picture bounding box.center) -- ++(-0.3cm,0)
                (path picture bounding box.center) -- ++(0,0.3cm)
                (path picture bounding box.center) -- ++(0,-0.3cm);
              },
          },
        otimes/.style={
            draw=black,
            very thick,
            circle,
            minimum size=8pt,
            inner sep=0pt,
            outer sep=0pt,
            path picture={
                \draw (path picture bounding box.center) -- ++(0.25cm,0.25cm)
                (path picture bounding box.center) -- ++(-0.25cm,-0.25cm)
                (path picture bounding box.center) -- ++(-0.25cm,0.25cm)
                (path picture bounding box.center) -- ++(0.25cm,-0.25cm);
              }
          },
        sigmoid/.style={
            draw=black,
            line width=1pt,
            circle,
            minimum size=8pt,
            inner sep=0pt,
            outer sep=0pt,
            path picture={
                \node at (path picture bounding box.center) {$\sigma$};
              }
          },
        swish/.style={
            draw=black,
            thick,
            line width=1pt,
            circle,
            minimum size=8pt,
            inner sep=0pt,
            outer sep=0pt,
            path picture={
                \draw[domain=-1.5:0, samples=50, variable=\x, blue, thick]
                plot ({\x}, {0});
                \draw[domain=0:1.5, samples=50, variable=\x, blue, thick]
                plot ({\x}, {\x});
              }
          },
        kgsa/.style={
            draw=black,
            very thick,
            fill=abc_color!50,
            minimum width=4cm,
            minimum height=0.8cm,
            rounded corners=3pt
          },
      }
      \begin{tikzpicture}
        \node[model] (model) at (0,0) {};
        \node[anchor=east,xshift=-2pt] at (model.west) (ntimes) {$N\times$};
        \node[below=12pt, minimum width=2.5cm] at (model.south) (input) {Inputs};
        \node[norm, anchor=south, yshift=20pt] at (model.south) (norm1) {Norm};
        \draw[layerlink] (input.north) -- (norm1.south);
        \node[tokenmixer, anchor=south, yshift=8pt] at (norm1.north) (gsa0) {\textsc{GSA}};
        \draw[normlink] (norm1.north) -- (gsa0.south);
        \node[oplus, anchor=south, yshift=4pt] at (gsa0.north) (oplus1) {};
        \node[norm, anchor=south, yshift=14pt] at (oplus1.north) (norm2) {Norm};
        \draw[normlink] (gsa0.north) -- (oplus1.south);
        \draw[layerlink] (oplus1.north) -- (norm2.south);
        \draw[residual] ([yshift=-10pt]norm1.south) -- ([xshift=-48pt,yshift=-10pt]norm1.south) -- ([xshift=-48pt]oplus1.center) -- (oplus1.center);
        \node[glu, anchor=south, yshift=8pt] at (norm2.north) (glu) {GLU};
        \draw[normlink] (norm2.north) -- (glu.south);
        \node[oplus, anchor=south, yshift=4pt] at (glu.north) (oplus2) {};
        \node[norm, anchor=south, yshift=10pt] at (model.north) (norm3) {Norm};
        \draw[normlink] (glu.north) -- (oplus2.south);
        \draw[residual] ([xshift=0pt,yshift=8pt]oplus1.center) -- ([xshift=-48pt,yshift=8pt]oplus1.center) -- ([xshift=-48pt]oplus2.center) -- (oplus2.center);
        \node[linear, anchor=south, yshift=5pt] at (norm3.north) (linear) {Linear};
        \draw[layerlink] (oplus2.north) -- (norm3.south);
        \draw[normlink] (norm3.north) -- (linear.south);
        \node[above=12pt] at (linear.north) (output) {Outputs};
        \draw[layerlink] (linear.north) -- (output.south);

        \node[gsa, anchor=west,xshift=30pt] (gsa) at (model.east)  {};

        \node[stacked, anchor=south, xshift=6pt, yshift=18pt, opacity=0.25] at ($(gsa.south west)!0.3!(gsa.south east)$) (vproj) {Linear};

        \node[stacked, anchor=south, yshift=20pt, opacity=0.5] at ($(gsa.south west)!0.3!(gsa.south east)$) (kproj) {Linear};
        \node[swish, xshift=0pt, yshift=48pt] at ($(gsa.south west)!0.3!(gsa.south east)$) (ksilu) {};
        \draw[normlink] (ksilu.south) -- (ksilu|-kproj.north);

        \node[stacked, anchor=south, xshift=-6pt, yshift=22pt] at ($(gsa.south west)!0.3!(gsa.south east)$) (qproj) {Linear};

        \node[fgate, anchor=south, yshift=20pt] at ($(gsa.south west)!0.7!(gsa.south east)$)  (fproj) {Linear};
        \node[sigmoid, yshift=48pt] at ($(gsa.south west)!0.7!(gsa.south east)$) (sigmoid) {};
        \draw[normlink] (fproj.north) -- (sigmoid.south);

        \node[below=12pt] at (gsa.south) (input) {\textcolor{white}{Inputs}};
        \draw[residual] (input.north) -- (gsa.south) |- ([xshift=-10pt,yshift=10pt]gsa.south) -| (kproj.south);
        \draw[residual] (input.north) -- (gsa.south) |- ([xshift=10pt,yshift=10pt]gsa.south) -| (fproj.south);

        \node[kgsa, yshift=-2pt] at (gsa.mid) (kernel) {Gated Slot Attention};
        \draw[layerlink] (ksilu.north) -- (ksilu|-kernel.south);
        \draw[qf] (sigmoid.north) -- (sigmoid.north|-kernel.south);
        \node[swish, anchor=south, yshift=5pt] at (kernel.north) (osilu) {};
        \draw[normlink] (kernel.north) -- (osilu.south);
        \node[norm, anchor=south, yshift=10pt] at (osilu.north) (norm4) {Norm};
        \draw[layerlink] (osilu.north) -- (norm4.south);
        \node[linear, anchor=south, yshift=5pt] at (norm4.north) (oproj) {Linear};
        \draw[normlink] (norm4.north) -- (oproj.south);
        \node[above=20pt,minimum width=2.5cm] at (oproj.north) (output) {\textcolor{white}{Outputs}};
        \draw[layerlink] (oproj.north) -- (output.south);
        \draw[modulelink] (gsa0.east) -| ([yshift=-10pt]$(model.east)!0.5!(gsa.west)$) |- (gsa.west);
      \end{tikzpicture}
    }
    \caption{\small
      The backbone of our proposed \textsc{GSA} models.
    }
    \label{fig:gsa}
  \end{minipage}

\end{figure}

The overall architecture of our proposed model, \textsc{GSA}, is shown in Figure~\ref{fig:gsa}. Following the Llama architecture~\citep{touvron-2023-llama}, we use a stack of $L$ \textsc{GSA} blocks, each comprising a \textsc{GSA} token mixing layer followed by a Gated Linear Unit (GLU) channel mixing layer~\citep{dauphin-2017-glu, hua-etal-2022-gau}.

We utilize the multi-head attention mechanism~\citep{vaswani-2017-attention} to capture different aspects of the input.
For each head $h$, the input to \textsc{GSA} token mixing is defined as
\begin{equation}
  \begin{aligned}
    \boldsymbol{q}_i^h,\boldsymbol{k}_i^h,\boldsymbol{v}_i^h & = \phi(\mathbf{W}_q^h \boldsymbol{x}_i),\phi(\mathbf{W}_k^h \boldsymbol{x}_i),\phi(\mathbf{W}_v^h \boldsymbol{x}_i) \\
  \end{aligned}
\end{equation}
where $\phi$ is the $\operatorname{Swish}$ activation following \citep{qin-2024-transnormerllm}.
The forget gate is obtained by a linear transformation followed by a sigmoid activation $\sigmoid$ with a damping factor $\tau$~\citep{yang-etal-2024-gla, sun-2024-yoco}: $\boldsymbol{\alpha}_i^h=\sigmoid(\mathbf{W}_{\alpha}^h \boldsymbol{x}_i)^{1/\tau}$,~\footnote{In practice we set $\tau=8$.} where the damping factor is to regulate the forget gate value to one, which has been shown to be crucial for long-term dependency modeling \citep{gu-2020-gating,qin-2023-hgrn}.
We feed them into a GSA layer to obtain outputs as described in Eq.~\ref{eq:gsa-2pass}:
\[
  \{\boldsymbol{o}_i^h\}_{i=1}^T = \operatorname{GSA}(\{\boldsymbol{q}_i^h, \boldsymbol{k}_i^h, \boldsymbol{v}_i^h, \boldsymbol{\alpha}_{i}^h\}_{i=1}^{T})
\]
Finally, we obtain output via
\begin{equation}
  \begin{aligned}
    \boldsymbol{y_i} & = \mathbf{W}_o\left(\operatorname{RMSNorm} \left(\operatorname{Swish}\left(\operatorname{Concat}\left(\boldsymbol{o}_i^1, \cdots, \boldsymbol{o}_i^H \right)\right)  \right)\right)
  \end{aligned}
\end{equation}

The total number of parameters for $\mathbf{W}_q, \mathbf{W}_k, \mathbf{W}_v,$ and $\mathbf{W}o$ is already $4d^2$, which is the same as in a single standard $\operatorname{softmax}$-attention layer.
To control the overall parameter count, we aim to keep the parameters for $\mathbf{W}_\alpha$, which amount to $dHm$, relatively small.
In practice, we set $m = 64$ to achieve a balance between efficiency and effectiveness (\S~\ref{sec:efficiency}).
One way to further manage the total parameter count is by reducing the number of heads.
In practice, we set $H = 4$, ensuring that $Hm \ll d$.
This keeps the total number of parameters approximately equal to $4d^2$.~\footnote{
  For instance, in a 1.3B model with $H \times m = 64 \times 4 = 256$ and $d = 2,048$, the total number of parameters amount to $4.125d^2$, introducing only a $0.125d^2$ overhead.
}

\section{Experiments}

\subsection{Language Modeling}

We perform moderate-scale language modeling experiments with 1.3B and 2.7B parameters on Slimpajama corpus \citep{cerebras-2023-slimpajama} for 100B tokens each.

We compare the performance of \textsc{GSA} against Llama Transformer architecture (i.e., Xfmr++~\citep{touvron-2023-llama} and recent subquadratic architectures including: Mamba~\citep{gu-2023-mamba}, RetNet~\citep{sun-2023-retnet}, GLA~\citep{yang-etal-2024-gla} and HGRN2~\citep{qin-2024-hgrn2}.
We refer readers to \S~\ref{sec:setup} for more details on baselines and other experimental setups.

\subsubsection{Results on commonsense reasoning tasks}

Following ~\cite{gu-2023-mamba,yang-etal-2024-gla}, we report the perplexities and zero-shot performance of commonsense reasoning tasks including
ARC$_e$ \& ARC$_c$ (ARC-easy, ARC-challenge)~\citep{clark-2018-arc};
Hella. (Hellaswag)~\citep{zellers-2019-hellaswag},
Lamb. (Lambada)~\citep{paperno-2019-lambada},
PIQA~\citep{bisk-2020-piqa},
Wiki. (Wikitext)~\citep{merity-2016-pointer}, and
Wino. (Winograde)~\citep{keisuke-2019-winogrande}. We note that these tasks are typically short in length and do not require in-context learning capabilities, thus they do not adequately reflect long-context modeling or in-context learning retrieval abilities. Nevertheless, as shown in Table~\ref{tab:main_results}, we found that \textsc{GSA} performs comparably to the recent strong model HGRN2 with an equally sized hidden state, while outperforming GLA and RetNet even with a smaller state size.
\begin{table*}[h!]
    \centering
    \small
    \renewcommand{\arraystretch}{1.1}
    \addtolength{\tabcolsep}{-1pt}
    \caption{
        The zero-shot results of 1.3B and 2.7B models evaluated by \texttt{lm-evaluation-harness} \citep{gao-2023-eval-harness}.
        $L$ denotes number of layer while $d$ denotes the model dimension.
    }
    \begin{tabular}{lrcc|ccccccc}
        \toprule
                     & \multirow{2}{*}{State                                                                                                                                                    size} & Lamb.                   & Wiki.                   & ARC$_e$       & ARC$_c$                 & Hella.                  & Lamb.         & PIQA          & Wino.         & \multirow{2}{*}{Avg.} \\
                     &                                                                                                                                                                                & ppl\rlap{$_\downarrow$} & ppl\rlap{$_\downarrow$} & acc           & acc\rlap{$_{\text{n}}$} & acc\rlap{$_{\text{n}}$} & acc           & acc           & acc           &                       \\
        \midrule
        \multicolumn{10}{l}{\emph{1.3B parameters with 100B training tokens, L=24, d=2,048}}                                                                                                                                                                                                                                                                                                          \\
        Xfmr++       & N/A                                                                                                                                                                            & 15.3                    & 17.1                    & 54.1          & 27.1                    & 49.3                    & 47.0          & 70.3          & \textbf{54.9} & 50.5                  \\
        Mamba        & $64\times Ld$                                                                                                                                                                  & 15.4                    & 17.3                    & 57.1          & 28.2                    & 50.3                    & 44.4          & 71.8          & 52.3          & 50.7                  \\
        RetNet       & $512\times Ld$                                                                                                                                                                 & 15.4                    & 17.3                    & 57.4          & 27.9                    & 50.3                    & 44.6          & 71.7          & 51.8          & 50.6                  \\
        GLA          & $256\times Ld$                                                                                                                                                                 & 15.4                    & 17.6                    & 55.4          & 27.7                    & 49.0                    & 46.4          & 69.9          & 54.0          & 50.4                  \\
        HGRN2        & $128\times Ld$                                                                                                                                                                 & \textbf{11.8}           & 16.9                    & \textbf{58.1} & 28.1                    & \textbf{51.8}           & \textbf{49.4} & 71.4          & 52.3          & \textbf{51.9}         \\
        \textsc{GSA} & $128\times Ld$                                                                                                                                                                 & 12.6                    & \textbf{16.7}           & \textbf{58.1} & \textbf{28.2}           & 51.0                    & 47.4          & \textbf{72.0} & 53.4          & 51.7
        \\
        \midrule
        \multicolumn{10}{l}{\emph{2.7B parameters with 100B training tokens, L=32, d=2,560}}                                                                                                                                                                                                                                                                                                          \\
        Xfmr++       & N/A                                                                                                                                                                            & 10.7                    & 15.2                    & 59.8          & 27.5                    & 54.2                    & 52.3          & 72.7          & \textbf{56.2} & 53.8                  \\
        Mamba        & $64\times Ld$                                                                                                                                                                  & 13.6                    & 15.9                    & 60.7          & 29.8                    & 53.9                    & 46.4          & 72.8          & 53.9          & 52.9                  \\
        RetNet       & $512\times Ld$                                                                                                                                                                 & 11.9                    & 15.8                    & 59.6          & 28.1                    & 54.0                    & 49.6          & 72.3          & 53.8          & 52.9                  \\
        GLA          & $256\times Ld$                                                                                                                                                                 & 12.4                    & 15.5                    & 59.2          & 29.9                    & 54.0                    & 50.4          & 71.7          & 55.7          & 53.5                  \\
        HGRN2        & $128\times Ld$                                                                                                                                                                 & \textbf{8.8}            & \textbf{14.6}           & 60.8          & 30.3                    & \textbf{58.7}           & \textbf{55.4} & 73.0          & 54.2          & \textbf{55.4}         \\
        \textsc{GSA} & $128\times Ld$                                                                                                                                                                 & 9.8                     & 14.8                    & \textbf{61.9} & \textbf{30.7}           & 57.0                    & 52.7          & \textbf{73.5} & 56.0          & 55.3                  \\
        \bottomrule
    \end{tabular}
    \label{tab:main_results}
\end{table*}

\subsubsection{Results on in-context recall-intensive tasks}\label{sec
} While subquadratic models can achieve comparable performance to (softmax-based) Transformers in language modeling and understanding tasks, their performance on recall-intensive tasks significantly lags behind Transformers and varies greatly across different subquadratic models, as observed in many recent studies \citep{arora-2024-simple, arora-2024-jrt, yang-etal-2024-gla,DBLP:journals/corr/abs-2406-06484}. Therefore, it is crucial to improve linear models on in-context recall-intensive tasks.

\begin{figure}[h!]
    \centering
    \begin{subfigure}[b]{0.33\textwidth}
        \centering
        \small
        \captionsetup{font=small}
        \resizebox{1\textwidth}{!}{
            \begin{tikzpicture}
                \begin{axis}[
                        trim axis left,
                        trim axis right,
                        ymajorgrids=true,
                        xmajorgrids=true,
                        tickwidth=0pt,
                        tick align=inside,
                        xlabel=Model dimension,
                        enlarge x limits=0.2,
                        width=6cm, height=5.5cm,
                        ymin=0, ymax=100,
                        symbolic x coords={64,128,256,512},
                        ytick={0,25,50,75,100},
                        yticklabels={0,25,50,75,100},
                        xlabel near ticks,
                        ylabel=Accuracy (\%),
                        ylabel style={at={(0.1,0.5)}},
                        axis line style={opacity=0},
                        legend style={
                                at={(1,0)},
                                anchor=south east,
                                legend cell align=left,
                                font=\small,
                            },
                    ]

                    \addplot[
                        mark=pentagon*,
                        draw=brickred!80,
                        thick,
                        mark options={
                                fill=brickred!70,
                                fill opacity=1.0,
                                solid
                            },
                        mark size=1.5pt,
                        opacity=1.0,
                    ] plot coordinates {
                            (64, 27.78)
                            (128,97.98)
                            (256,99.06)
                            (512,99.00)
                        };
                    \addlegendentry{\scriptsize GSA}

                    \addplot[
                        thick,
                        dashdotdotted,
                        mark=star,
                        mark size=2pt,
                        mark options={scale=1},
                        color=darkcyan
                    ] plot coordinates {
                            (64, 0.01)
                            (128,99.02)
                            (256,99.05)
                            (512,100)
                        };
                    \addlegendentry{\scriptsize Mamba}

                    \addplot[
                        densely dashdotted,
                        mark=square*,
                        draw=blue!60, thick,
                        mark size=1pt,
                        mark options={fill=blue!60, fill opacity=1.0, solid},
                        opacity=1.0,
                    ] plot coordinates {
                            (64, 04.12)
                            (128,85.05)
                            (256,98.72)
                            (512,99.11)
                        };
                    \addlegendentry{\scriptsize GLA}

                    \addplot[
                        dashed,
                        mark=10-pointed star,
                        mark size=2pt,
                        draw=orange!70,
                        thick,
                    ] plot coordinates {
                            (64, 00.82)
                            (128,13.92)
                            (256,97.04)
                            (512,99.04)
                        };
                    \addlegendentry{\scriptsize RetNet}

                    \addplot[
                        thick,
                        densely dashdotted,
                        mark=x,
                        mark size=2pt,
                    ] plot coordinates {
                            (64,0.01)
                            (128,0.01)
                            (256,99)
                            (512,99)
                        };
                    \addlegendentry{\scriptsize HGRN2}

                    \draw [semithick, black] (axis description cs:0,0) -- (axis description cs:1,0); %
                    \draw [semithick, black] (axis description cs:0,1) -- (axis description cs:1,1); %

                \end{axis}

            \end{tikzpicture}
        }
        \caption{
            Results on the synthetic MQAR task.
            We adopt the most challenging settings in \cite{arora-2023-zoology}, utilizing a sequence length of 512 and 64 key-value pairs.
            Xfmr++ with standard attention achieves near-perfect results in this settings and is thus omitted for brevity.
        }
        \label{fig:mqar}
    \end{subfigure}%
    \hfill
    \begin{subfigure}[b]{0.64\textwidth}
        \renewcommand{\arraystretch}{1.1}
        \caption{
            \small{
                Results on the recall-intensive tasks used in \cite{arora-2024-jrt}.
                We truncate the input to a maximum of 2K tokens.
            }
        }
        \centering
        \footnotesize
        \addtolength{\tabcolsep}{-4.5pt}
        \begin{tabular}{lrccccccc}
            \toprule
                                      & State size     & FDA                     & SWDE          & SQuAD         & NQ            & TriviaQA      & Drop          & Avg.          \\
            \midrule
            \multicolumn{5}{l}{\emph{1.3B params / 100B tokens, L=24, d=2048}}                                                                                                   \\
            \rowcolor{gray!15} Xfmr++ & N/A            & 46.0                    & 29.2          & 41.0          & 24.8          & 58.8          & 21.3          & 36.9          \\
            Mamba                     & $64\times Ld$  & 13.9                    & 25.4          & 33.2          & 18.5          & 53.5          & \textbf{21.7} & 27.7          \\
            RetNet                    & $512\times Ld$ & 21.2                    & 27.2          & 34.0          & 15.5          & 52.7          & 20.0          & 28.4          \\
            GLA                       & $256\times Ld$ & \textbf{26.7}           & \textbf{30.6} & 34.8          & 21.5          & 56.0          & 19.1          & 31.4          \\
            HGRN2                     & $128\times Ld$ & \textcolor{white}{0}9.9 & 23.1          & 32.0          & 16.4          & 55.2          & 19.1          & 25.9          \\
            GSA                       & $128\times Ld$ & 23.6                    & 29.8          & \textbf{36.0} & \textbf{23.2} & \textbf{57.0} & 20.9          & \textbf{31.8} \\
            \midrule
            \multicolumn{5}{l}{\emph{2.7B params / 100B tokens, L=32, d=2560}}                                                                                                   \\
            \rowcolor{gray!15} Xfmr++ & N/A            & 62.3                    & 30.9          & 44.3          & 29.3          & 61.8          & 21.4          & 41.7          \\
            Mamba                     & $64\times Ld$  & 21.5                    & 26.7          & 34.2          & 21.2          & 57.0          & \textbf{22.2} & 30.5          \\
            RetNet                    & $512\times Ld$ & 24.1                    & 26.1          & 36.4          & 20.4          & 57.3          & 21.8          & 31.0          \\
            GLA                       & $256\times Ld$ & 30.3                    & \textbf{35.5} & 36.8          & 23.3          & 58.2          & 21.8          & 34.3          \\
            HGRN2                     & $128\times Ld$ & 15.0                    & 29.9          & 35.1          & 17.0          & 59.8          & 20.0          & 29.5          \\
            GSA                       & $128\times Ld$ & \textbf{39.1}           & 33.5          & \textbf{39.0} & \textbf{26.9} & \textbf{60.8} & 19.9          & \textbf{36.5} \\
            \bottomrule
        \end{tabular}
        \label{tab:recall}
    \end{subfigure}
\end{figure}

\paragraph{MQAR.} We first present the results on the multi-query associative recall (MQAR) task \cite{arora-2023-zoology}, a diagnostic synthetic task that requires models to retrieve multiple associative key-value pairs from the context. This task has been shown to strongly correlate with language modeling performance \citep{arora-2023-zoology}. The results in Table~\ref{fig:mqar}
validate the effectiveness of \textsc{GSA}.

\begin{wraptable}[]{r}{0.4\textwidth}
    \vspace{-1\normalbaselineskip}
    \caption{\small
        Ablation study results for 340M models trained on 10B Slimpajama tokens. 
    }
    \centering
    \small
    \renewcommand{\arraystretch}{1.1}
    \addtolength{\tabcolsep}{-2pt}
    \resizebox{.4\textwidth}{!}{
    \begin{tabular}{lc}
        \toprule
                                        & PPL  (${\downarrow}$) \\
        \midrule
        \textsc{GSA} w/ 64 slots        & 13.51                 \\
        \midrule
        \emph{Ablations on gating mechanism}                    \\
        \quad w/o decay (i.e., ABC)     & 16.94                 \\
        \quad w/ data-independent decay & 15.83                 \\
        \midrule
        \emph{Ablations on non-linearity}                       \\
        \quad $- \operatorname{softmax}$                 & 14.03                 \\
        \quad $- \operatorname{softmax} + \operatorname{Swish}$         & 13.71                 \\
        \quad $- \operatorname{softmax} + \operatorname{ReLU}$          & 13.69                 \\
        \quad $- \operatorname{softmax} + \operatorname{ReLU}^2$      & 13.95                 \\
        \midrule
        \emph{Ablations on slot size}                           \\
        \quad w/ 32 slots               & 13.74                 \\
        \quad w/ 128 slots              & \textbf{13.46}        \\
        \bottomrule
    \end{tabular}
    }
    \label{tab:ablations}

\vspace{-10mm}
\end{wraptable}

\paragraph{Real-world tasks.} Next, we evaluate the zero-shot in-context learning performance on recall-intensive tasks, as used in \citet{arora-2024-jrt}.\footnote{Since our pretrained models are neither instruction-tuned nor instruction-aligned, following \citet{arora-2024-jrt}, we use their Cloze Completion Formatting prompts for evaluation. It is noteworthy that results for certain tasks may differ significantly from those obtained using \texttt{lm-evaluation-harness} \citep{gao-2023-eval-harness} due to variations in prompt templates.} Specifically, we assess information retrieval on FDA \citep{wu-2021-fda} and SWDE \citep{lockard-2019-openceres}, which are designed to evaluate retrieval from in-context passages scraped from HTML/PDFs. We also evaluate question answering on SQuAD \citep{rajpurkar-2018-know}, NQ \citep{kwiatkowski-etal-2019-natural}, TriviaQA \citep{joshi-etal-2017-triviaqa}, and Drop \citep{dua-etal-2019-drop}, where models must ground their answers in in-context documents.

As shown in Table \ref{tab:recall}, Xfmr++ achieves the best average performance, as expected. Meanwhile, \textsc{GSA} outperforms all other subquadratic baseline models by a notable margin without requiring a larger state size. We believe this advantage stems from \textsc{GSA}'s context-aware memory readout mechanism (as discussed in \S\ref{sec:abc}) and its forgetting mechanism (i.e., the gating mechanism), enabling it to manipulate finite-sized memory more effectively.

\subsubsection{Ablation}

Table~\ref{tab:ablations} presents the results of our ablation studies.  Our findings indicate that:
(i) the inclusion of the gating mechanism in GSA is crucial for improving language modeling perplexity;
(ii) applying $\operatorname{softmax}$ non-linearities after the first recurrent pass is beneficial; and
(iii) using 64 slots strikes an optimal balance between performance and efficiency. \footnote{
  Empirically, we found that 32, 64, and 128 slots result in training throughputs of 46.7K, 44.1K, and 37.1K tokens/s, respectively, under the settings described in the next section. Given the marginal improvement when increasing the slot size from 64 to 128, along with the significant slowdown in training, we chose 64 slots.
}

\subsubsection{Efficiency}
\label{sec:efficiency}

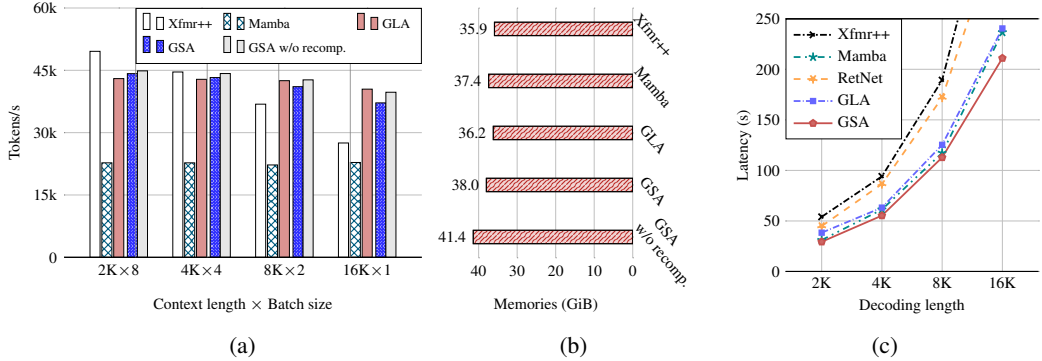
\begin{figure}[h!]
    \centering
    \begin{subfigure}[b]{0.35\textwidth}
        \centering
        \small
        \captionsetup{font=small}
        \centering
        \resizebox{\textwidth}{!}{
            \begin{tikzpicture}[
                    trim axis left,
                    trim axis right,
                    my grid/.style={thin, gray, opacity=0.8},
                ]
                \pgfplotstableread{speed.dat}\loadedtable
                \begin{axis}[
                        footnotesize,
                        ymajorgrids,
                        yminorgrids,
                        tick align=inside,
                        axis line style={opacity=0},
                        tickwidth=0pt,
                        width=7.2cm, height=5.5cm,
                        enlarge x limits=0.22,
                        ybar=2*\pgflinewidth,
                        bar width=4.5pt,
                        legend style={at={(1, 1)},
                                anchor=north east, legend columns=3,
                                /tikz/every even column/.append style={column sep=0.1cm},
                                font=\tiny},
                        legend cell align={left},
                        symbolic x coords={
                                2K,
                                4K,
                                8K,
                                16K,
                                M,
                            }, %
                        xtick=data,
                        xticklabels={
                                2K$\times$8,
                                4K$\times$4,
                                8K$\times$2,
                                16K$\times$1,
                                Memory
                            },
                        x tick label style={font=\scriptsize, align=center}, %
                        y tick label style={font=\scriptsize, align=center},
                        xtick distance=1,
                        xlabel={\scriptsize Context length $\times$ Batch size},
                        xlabel style={font=\scriptsize,at={(0.5,0)}},
                        scaled ticks=false,
                        ymin=0, ymax=60000,
                        ytick={0,15000,30000,45000,60000},
                        yticklabels={0, 15k, 30k, 45k,60k},
                        ylabel={\scriptsize Tokens/s},
                        ylabel style={font=\scriptsize,at={(0.08,0.5)}},
                        axis y line*=left
                    ]
                    \addplot[draw=black, fill=white] table[x=category,y=Xfmr++] {\loadedtable};
                    \addlegendentry{\tiny Xfmr++}
                    \addplot[draw=black, pattern=crosshatch, pattern color=midnightblue] table[x=category,y=Mamba] {\loadedtable};
                    \addlegendentry{\tiny Mamba}
                    \addplot[draw=black, fill=brickred!50] table[x=category,y=GLA] {\loadedtable};
                    \addlegendentry{\tiny GLA}
                    \addplot[draw=black, pattern={Lines[angle=45,distance={1.2pt/sqrt(2)},line width=.8pt]},pattern color=blue!80] table[x=category,y=GSA] {\loadedtable};
                    \addlegendentry{\tiny GSA}
                    \addplot[draw=black, fill=gray!20] table[x=category,y=GSA1] {\loadedtable};
                    \addlegendentry{\tiny GSA w/o recomp.}
                    \draw [semithick, black] (axis description cs:0,0) -- (axis description cs:1,0); %
                    \draw [my grid] (axis description cs:0.2,0) -- (axis description cs:0.2,1);
                    \draw [my grid] (axis description cs:0.4,0) -- (axis description cs:0.4,1);
                    \draw [my grid] (axis description cs:0.6,0) -- (axis description cs:0.6,1);
                    \draw [my grid] (axis description cs:0.8,0) -- (axis description cs:0.8,1);
                \end{axis}
            \end{tikzpicture}
        }
        \caption{
        }
        \label{fig:speed}
    \end{subfigure}%
    \begin{subfigure}[b]{0.28\textwidth}
        \centering
        \small
        \resizebox{\textwidth}{!}{
            \begin{tikzpicture}
                \begin{axis}[
                        footnotesize,
                        xmajorgrids,
                        xminorgrids,
                        tick align=inside,
                        axis line style={opacity=0},
                        tickwidth=0pt,
                        xbar,  %
                        width=4.3cm, height=5.5cm,  %
                        bar width=6pt,
                        xbar=2*\pgflinewidth,
                        symbolic y coords={{GSA w/ naive impl.}, {GSA \\\quad\quad w/o recomp.}, GSA, GLA, Mamba, Xfmr++},  %
                        xtick=data,
                        xmin=0, xmax=45,
                        xtick={0,10,20,30,40},
                        xticklabels={0,10,20,30,40},
                        x tick label style={font=\scriptsize, align=center}, %
                        y tick label style={font=\scriptsize, align=center},
                        ytick distance=1,
                        nodes near coords,
                        nodes near coords style={font=\scriptsize,align=center,anchor=east, text=black},
                        point meta=explicit symbolic,
                        xlabel={\scriptsize Memories (GiB)},  %
                        x dir=reverse,  %
                        yticklabel pos=right,  %
                        y tick label style={
                                at={(current axis.east)},
                                anchor=base,
                                rotate=-50,
                                xshift=8pt,
                                yshift=3pt
                            },
                        xticklabel style={rotate=0},
                    ]

                    \addplot[
                        draw=black,
                        fill=brickred!20,
                        postaction={
                                pattern={Lines[angle=45,distance={1.8pt/sqrt(2)},line width=.4pt]},
                                pattern color=brickred,
                            }
                    ] table [
                    x=x,
                    y=y,
                    meta=label,
                    row sep=crcr
                    ] {
                    x       y                             label\\
                    35.9   Xfmr++                         {35.9}\\
                    37.4   Mamba                          {37.4}\\
                    36.2   GLA                            {36.2}\\
                    38.0   GSA                            {38.0}\\
                    41.4   {GSA \\\quad\quad w/o recomp.}            {41.4}\\
                    };
                \end{axis}
            \end{tikzpicture}
        }
        \caption{}
        \label{fig:memory}
    \end{subfigure}
    \begin{subfigure}[b]{0.3\textwidth}
        \centering
        \scriptsize
        \resizebox{\textwidth}{!}{
            \begin{tikzpicture}
                \begin{axis}[
                        trim axis left,
                        trim axis right,
                        ymajorgrids=true,
                        xmajorgrids=true,
                        tickwidth=0pt,
                        tick align=inside,
                        xlabel=Decoding length,
                        enlarge x limits=0.2,
                        width=5.5cm, height=5.5cm,
                        ymin=0, ymax=250,
                        symbolic x coords={2K,4K,8K,16K},
                        ytick={0,50,100,150,200,250},
                        yticklabels={0,50,100,150,200,250},
                        xlabel near ticks,
                        ylabel=Latency (s),
                        ylabel style={at={(0.15,0.5)}},
                        axis line style={opacity=0},
                        legend style={
                                at={(0,1)},
                                anchor=north west,
                                legend cell align=left,
                                font=\small,
                            },
                    ]

                    \addplot[
                        thick,
                        densely dashdotted,
                        mark=x,
                        mark size=2pt,
                    ] plot coordinates {
                            (2K, 54.1)
                            (4K, 94.1)
                            (8K, 189.3)
                            (16K,418.3)
                        };
                    \addlegendentry{\scriptsize Xfmr++}

                    \addplot[
                        thick,
                        dashdotdotted,
                        mark=star,
                        mark size=2pt,
                        mark options={scale=1},
                        color=darkcyan
                    ] plot coordinates {
                            (2K, 31.1)
                            (4K,61.3 )
                            (8K, 117.3 )
                            (16K,236.6  )
                        };
                    \addlegendentry{\scriptsize Mamba}

                    \addplot[
                        dashed,
                        mark=10-pointed star,
                        mark size=2pt,
                        draw=orange!70,
                        thick,
                    ] plot coordinates {
                            (2K,44.9)
                            (4K,86.5 )
                            (8K,172.0)
                            (16K,347.8)
                        };
                    \addlegendentry{\scriptsize RetNet}

                    \addplot[
                        densely dashdotted,
                        mark=square*,
                        draw=blue!60, thick,
                        mark size=1pt,
                        mark options={fill=blue!60, fill opacity=1.0, solid},
                        opacity=1.0,
                    ] plot coordinates {
                            (2K, 38.4)
                            (4K,63.2)
                            (8K,125.3)
                            (16K,240.4)
                        };
                    \addlegendentry{\scriptsize GLA}

                    \addplot[
                        mark=pentagon*,
                        draw=brickred!80,
                        thick,
                        mark options={
                                fill=brickred!70,
                                fill opacity=1.0,
                                solid
                            },
                        mark size=1.5pt,
                        opacity=1.0,
                    ] plot coordinates {
                            (2K, 29.4)
                            (4K,55.3)
                            (8K,112.8)
                            (16K,211.0)
                        };
                    \addlegendentry{\scriptsize GSA}

                    \draw [semithick, black] (axis description cs:0,0) -- (axis description cs:1,0); %
                    \draw [semithick, black] (axis description cs:0,1) -- (axis description cs:1,1); %

                \end{axis}

            \end{tikzpicture}
        }
        \caption{}
        \label{fig:inference}
    \end{subfigure}

    \caption{
        \subref{fig:speed} Training throughput of various 1.3B models on a single H800 GPU, with a fixed batch size containing 16K tokens. ``\textsc{GSA} w/o recomp.'' indicates the use of the \textsc{GSA} kernel without hidden state recomputation during the backward pass.
        \subref{fig:memory} Memory footprint (in GiB) of each 1.3B model during training with a batch size containing 16K tokens.
        \subref{fig:inference} Inference latency (in seconds) of each 1.3B model on a single H800 GPU with 2K prefix tokens and a batch size of 1.
    }
\end{figure}

Fig.~\ref{fig:speed} illustrates the training throughput for four models on a single H800 GPU\footnote{We utilize the training throughput benchmark scripts provided by FLA \cite{yang-2024-fla} for our measurements.}.
To optimize memory usage, we employ the technique of recomputing the recurrent hidden state during the backward pass, as done in FLA \citep{yang-2024-fla} and Mamba2 \cite{mamba2}.
This approach results in reduced memory consumption (Fig.~\ref{fig:memory}) at the cost of slightly lower training throughputs (Fig.~\ref{fig:speed}).

Despite requiring two GLA passes, GSA maintains comparable training throughputs to GLA due to its reduced state size. Since inference is primarily memory-bound, inference speed highly correlates with state size. As a result, GSA, with its smaller state size compared to RetNet and GLA, achieves faster inference speeds, as shown in Figure~\ref{fig:inference}.

\begin{table*}[h!]
    \centering
    \small
    \renewcommand{\arraystretch}{1.1}
    \addtolength{\tabcolsep}{-3pt}
    \caption{\small
        Performance comparison across various 7B models.
        $^\clubsuit$ denotes models using $\operatorname{softmax}$-attention.
        $^\dagger$ denotes our results.
    }
    \begin{tabular}{lrrccccccccccc}
        \toprule
                                   & Size & Tokens & ARC$_e$       & ARC$_c$       & Hella.        & PIQA          & Wino.         & NQ            & TriviaQA      & BBH                     & MMLU          & \multirow{2}{*}{Avg.} \\
        Shot(s)                    &      &        & 0             & 0             & 0             & 0             & 0             & 5             & 5             & 3                       & 5                                     \\[1pt]
        \midrule

        \multicolumn{13}{l}{\emph{Models trained from scratch (for reference)}}                                                                                                                                                      \\[1pt]
        RWKV6                      & 7B   & 1.4T   & 73.6          & 44.0          & 75.2          & 78.4          & 68.5          & 20.9          & 59.5          & 23.4                    & 43.9          & 54.1                  \\
        Mamba                      & 7B   & 1.2T   & 77.6          & 46.8          & 77.8          & 81.0          & 72.3          & 25.4          & 66.2          & 21.5                    & 33.2          & 55.7                  \\
        Llama2\rlap{$^\clubsuit$}  & 7B   & 2T     & 76.4          & 46.2          & 76.0          & 78.0          & 69.2          & 26.0          & 64.2          & 39.1                    & 45.5          & 57.8                  \\
        Gemma\rlap{$^\clubsuit$}   & 7B   & 6T     & 81.5          & 53.2          & 80.5          & 79.8          & 74.0          & 24.3          & 63.7          & 58.9                    & 63.2          & 64.3                  \\
        Mistral\rlap{$^\clubsuit$} & 7B   & $?$    & 80.8          & 54.0          & 81.1          & 80.6          & 74.0          & 29.7          & 70.3          & 56.5                    & 62.4          & 65.5                  \\[1pt]
        \multicolumn{13}{l}{\emph{Models finetuned from Mistral 7B}}                                                                                                                                                                 \\[1pt]
        SUPRA                      & 7B   & +20B   & 74.6          & 42.3          & 74.8          & \textbf{80.1} & 67.4          & -             & -             & -                       & 28.0          & -                     \\
        RetNet\rlap{$^\dagger$}    & 7B   & +20B   & 73.3          & 39.9          & 72.9          & 77.8          & 66.1          & 16.2          & 43.0          & \textcolor{white}{0}8.7 & 26.1          & 47.1                  \\
        GLA\rlap{$^\dagger$}       & 7B   & +20B   & 74.6          & \textbf{44.0} & 75.9          & 79.2          & 69.5          & 22.2          & 57.8          & 20.8                    & 28.4          & 52.5                  \\
        GSA\rlap{$^\dagger$}       & 7B   & +20B   & \textbf{75.9} & 43.9          & \textbf{76.5} & 78.7          & \textbf{70.1} & \textbf{23.4} & \textbf{60.7} & \textbf{23.5}           & \textbf{32.4} & \textbf{53.9}         \\[3pt]
        SUPRA                      & 7B   & +100B  & \textbf{76.0} & 45.7          & 77.1          & \textbf{79.9} & 70.3          & 24.7          & 60.4          & 19.8                    & 34.1          & 54.2                  \\
        GSA\rlap{$^\dagger$}       & 7B   & +100B  & \textbf{76.0} & \textbf{46.9} & \textbf{77.9} & 78.9          & \textbf{72.6} & \textbf{26.9} & \textbf{65.8} & \textbf{29.3}           & \textbf{38.1} & \textbf{56.9}         \\
        \bottomrule
    \end{tabular}
    \label{tab:cpt}
\end{table*}

\subsection{Finetuning Pretrained Transformers to RNNs}

\label{sec:continual}
The concept of finetuning pretrained Transformers to linear Transformers for recurrent inference was first introduced in T2R \citep{kasai-etal-2021-finetuning}.
This approach uses pretrained language model weights to initialize all parameters, leveraging the similarity between linear attention and $\operatorname{softmax}$ attention, and finetunes all parameters, significantly reducing the total training time compared to training from scratch.
\citet{kasai-etal-2021-finetuning} also introduced a \emph{parametric} feature map, implemented as a learnable MLP layer followed by $\operatorname{ReLU}$, applied after the query/key projections.
SUPRA, a follow-up to T2R, found that the original T2R approach did not perform well in the era of LLMs, and highlighted the importance of output normalization and a decay mechanism—adopted from RetNet \citep{sun-2023-retnet}—as critical for finetuning performance.
As a result, SUPRA essentially combines T2R and RetNet by finetuning pretrained Transformers into a RetNet architecture, though it excludes the $\operatorname{Swish}$ output gate.

\paragraph{Settings.}
In our preliminary experiments, we found that the learnable MLP layer was unnecessary and could be merged into the query and key projections, similar to the approach in \citet{peng-etal-2022-abc}.
We finetuned the pretrained Transformer Mistral 7B \citep{jiang-2023-mistral} to RetNet, as well as to GLA and GSA models.
Following SUPRA, we add $\operatorname{ReLU}$ as the feature map activation for RetNet and GLA, which originally used an identity feature map without activation~\footnote{However, SUPRA reported poor performance with this strategy due to a significant discrepancy between training and finetuning, where an identity map can lead to negative attention scores, a pattern unseen in pretrained Transformers due to the nonnegativity of $\operatorname{softmax}$.}, and also excluded the $\operatorname{Swish}$ output gate.
For RetNet, there were no additional parameters; for GLA, the low-rank forget gate, and for GSA, the $W_{\alpha}$ matrix are trainable parameters, though both are small in parameter count and negligible in terms of the total model size.
We set the peak learning rate to $3\times 10^{-5}$ with 1K steps of linear warmup following SUPRA.
The training  length was set to 2K tokens, with a batch size of 2M tokens.
For convenience, we trained on the SlimPajama corpus, while SUPRA used RefineWeb \cite{zhang-2023-refinedweb}, a higher-quality corpus.
We leave the use of RefineWeb for future work.

\paragraph{Main results.}
Following \citet{jiang-2023-mistral,touvron-2023-llama2}, we evaluated the models on commonsense reasoning tasks: ARC$_e$ and ARC$_c$ \citep{clark-2018-arc}, Hellaswag \citep{zellers-2019-hellaswag}, PIQA \citep{bisk-2020-piqa}, and Winogrande \citep{keisuke-2019-winogrande}; world knowledge tasks: NQ \citep{kwiatkowski-etal-2019-natural} and TriviaQA \citep{joshi-etal-2017-triviaqa}; and popular aggregated benchmarks: MMLU \citep{hendrycks-2021-mmlu} and BBH \citep{suzgun-etal-2023-challenging}.
Results are shown in Table \ref{tab:cpt}.
We observed a clear advantage in finetuning Mistral to GSA compared to GLA or RetNet, confirming our intuition that preserving $\operatorname{softmax}$ is beneficial in T2R settings.
When trained with 100B tokens, Mistral-to-GSA outperforms RWKV6 and Mamba on average, even though those models were trained on over 1T tokens, thereby reducing the required training data size.

\begin{wraptable}[12]{r}{0.5\textwidth}
    \vspace{-5mm}
    \centering
    \small
    \addtolength{\tabcolsep}{-3.5pt}
    \renewcommand{\arraystretch}{1.1}
    \caption{
        Long-context performance  comparison.
    }
    \begin{tabular}{lcccc}
        \toprule
                                   & Qasper                  & NarrativeQA             & QuALITY       & QMSum                   \\[1pt]
        \midrule
        \multicolumn{5}{l}{\emph{Models trained from scratch (for reference)}}                                                   \\[1pt]
        RWKV6                      & \textcolor{white}{0}9.2 & 14.4                    & 30.8          & \textcolor{white}{0}1.1 \\
        Mamba                      & \textcolor{white}{0}5.6 & 27.9                    & 27.5          & \textcolor{white}{0}0.8 \\
        Mistral\rlap{$^\clubsuit$} & 25.8                    & 25.1                    & 38.0          & \textcolor{white}{0}5.0 \\[1pt]
        \multicolumn{5}{l}{\emph{Models finetuned from Mistral 7B on 20B tokens}}                                                \\[1pt]
        RetNet                     & 11.1                    & \textcolor{white}{0}0.0 & 26.2          & \textcolor{white}{0}0.0 \\
        GLA                        & 18.4                    & 17.2                    & 30.9          & \textcolor{white}{0}9.0 \\
        GSA                        & \textbf{18.8}           & \textbf{19.2}           & \textbf{32.0} & \textbf{10.0}           \\
        \bottomrule
    \end{tabular}
    \label{tab:long_eval}
\end{wraptable}
\paragraph{Long-context ability evaluation.}
Following \citet{xiong-2023-llamalong}, we evaluated the models on long-sequence tasks, including Qasper \citep{dasigi-etal-2021-dataset}, NarrativeQA \citep{kocisky-etal-2018-narrativeqa}, QuALITY \citep{pang-2022-quality}, and QMSum \citep{zhong-etal-2021-qmsum}. For each task, the input was truncated to 16K tokens, which is 8$\times$ the training length.

The results are shown in Table~\ref{tab:long_eval}. Notably, \textsc{GSA} consistently outperforms other subquadratic models across all four tasks. We attribute this to the same factors observed in in-context recall-intensive task settings. Interestingly, Mistral-to-GSA also demonstrates overall better performance compared to RWKV6 and Mamba, which were trained from scratch on $>$1T token.

\section{Related works}

\paragraph{Matrix-valued linear RNNs with hardware-efficient training.}
Traditional RNNs (e.g., LSTM \cite{hochreiter-1997-lstm}, GRU \cite{cho-etal-2014-gru}) maintain 1-dimensional hidden states, which are often too small to capture sufficient information. Recent work emphasizes the importance of expanding the size of recurrent states \citep{gu-2023-mamba, qin-2024-hgrn2, yang-etal-2024-gla, sun-2023-retnet, peng-2024-eagle, mamba2}. However, naive state expansion dramatically increases FLOPs and I/O costs, making training impractical. To address this, Mamba introduces an I/O-aware approach, reducing I/O costs by materializing parameters and hidden states only on SRAM (instead of HBM).
However, Mamba's recurrence cannot be expressed in matmul form, leading to two key issues: (i) high FLOP count cannot be optimized via tensor cores (the GPU's fast matmul unit), resulting in slower runtimes; and (ii) the recurrent hidden states cannot be compactly represented and must be materialized on SRAM during backpropagation, limiting the recurrent state size due to SRAM constraints.

Mamba2 \cite{mamba2} addresses these limitations by adopting a linear attention \cite{katharopoulos-2020-transformers}-like approach that enables hardware-efficient training. Linear attention expands the state using outer products, allowing for both parallel attention-style computation and recurrent inference (also known as state-space duality in Mamba2). The chunkwise algorithm interpolates between parallel and recurrent forms, enabling hardware-efficient, linear-time training \cite{hua-etal-2022-gau, sun-2023-retnet, yang-etal-2024-gla}. However, vanilla linear attention underperforms $\operatorname{softmax}$ attention in various tasks. Recent research has explored incorporating various decay or gating mechanisms to enhance model expressiveness and performance while maintaining matmul-based parallelism and chunkwise training. These include head-wise data-independent decay \cite{sun-2023-retnet, qin-2024-transnormerllm}; head-wise data-dependent decay \cite{peng-2021-rfa, mamba2, beck-2024-xlstm, sun-2024-yoco}; and channel-wise data-dependent decay \cite{yang-etal-2024-gla, mao-2022-fine,  katsch-2024-gateloop, qin-2024-hgrn2, peng-2024-eagle}. \textsc{GSA} leverages two-pass gated linear attention to further enhance capacity while allowing hardware-efficient training.

\paragraph{Fast weight RNNs.}

Fast weight programming \citep{schmidhuber1992learning}, a classical concept intensively investigated in deep learning \citep{ba-2016-using,zhang-2017-learning,schlag-2017-gated,schlag-2021-learning,munkhdalai-2019-metalearned,schlag-2021-deltanet,Irie-2021-going, Irie-2022-ams,mao-2022-fine}, has been shown to be closely related to (linear) Transformers \citep{schlag-2021-deltanet}.
The core idea involves using a slow network to produce rapid context-dependent weight modifications for the fast network.
In linear attention, the fast network is a single-layer FFN with weight matrix $\mathbf{S}_t$ (Eq.~\ref{eq:la}), while the slow networks are the query/key/value projections.

Linear attention is known to suffer from limited memory capacity \citep{schlag-2021-deltanet}, potentially due to the constraints of a single-layer FFN without a large representation.
In contrast, ABC and \textsc{GSA} can be viewed as implementing a two-layer fast FFN with either additive update rule or gated update rule \cite{schlag-2017-gated,mao-2022-fine}, where the weight matrices are $\widetilde{\mathbf{K}}_{t}$ and $\widetilde{\mathbf{V}}_{t}$ connected by the $\operatorname{softmax}$ activation function (Eq.~\ref{eq:abc_recurrent} and Eq.~\ref{eq:gsa}).
This structure resembles DeltaMLP \cite{Irie-2021-going}, which uses a delta update rule \citep{widrow-1988-adaptive, schlag-2021-deltanet, yang-2024-parallelizing} and a multi-layer (potentially beyond two layers) fast FFN. The greater capacity of a two-layer FFN compared to a similarly sized single-layer FFN could explain why \textsc{GSA} requires a smaller state size to achieve similar or even better performance, especially in long sequence and recall-intensive tasks.

\paragraph{Finetuning Transformers to RNNs.}
As discussed, this paradigm could significantly reduce the training cost for large-scale recurrent language models. The idea of distilling Transformers to RNNs to improve inference efficiency can be traced back to \citet{gerstenberger-2020-icassp-domain}. In the following, we briefly introduce some recent works that complement those already mentioned in \S\ref{sec:continual}
. \citet{zhang-2024-hedgehog} highlight the desirable properties of $\operatorname{softmax}$, such as attention spikiness and dot-product monotonicity, and employ a learnable MLP layer to approximate $\operatorname{softmax}$ behavior using logit distillation loss (while freezing other parameters). \citet{chen-2024-dijiang} introduce DiJiang, an effective method for approximating attention distributions using the Discrete Cosine Transform (DCT) to enable frequency-domain kernelization, leading to faster feature mapping. \citet{bick-2024-transformertossms} propose a multi-stage distillation approach, aligning attention distributions (similar to Hedgehog \cite{zhang-2024-hedgehog}), hidden states, and output logits to transfer knowledge from a pretrained Transformer teacher to a student Mamba model. \citet{wang-2024-mamballama} distill Transformer-based LLMs into hybrid Mamba-Attention architectures in the spirit of \citet{samba,jamba,mamba2hybrid}. However, they freeze the FFN weights, while \citet{choi-2024-cross} suggest that it might be more effective to unfreeze them. In this work, we highlight the importance of the $\operatorname{softmax}$ operator, as discussed in \citet{zhang-2024-hedgehog}, except that \textsc{GSA} directly incorporates $\operatorname{softmax}$, while \citet{zhang-2024-hedgehog} learns a feature map to mimic $\operatorname{softmax}$, without actually including any $\operatorname{softmax}$ operator in the resulting model.

\section{Limitations and future work}

Due to the relatively small scale of our pretrained models (compared to large-scale models trained on trillions of tokens), we did not report any results on long-context tasks, as the performance would all be poor.
However, we believe Table~\ref{tab:long_eval} provides positive indications of GSA's long-context capabilities, and training on a larger token horizon and with larger models would address this.
For copy-oriented tasks, we observed negative results on the Phonebook Lookup \cite{jelassi-2024-repeat} and Needle-In-Haystack evaluations compared to Transformers, revealing the fundamental limitations of linear recurrent models in handling ``precise local token shifts and comparison'', as discussed in \citet{arora-2024-simple}. Nonetheless, we expect this limitation could be significantly mitigated by pretraining a hybrid GSA-attention model, as recently explored \cite{arora-2024-simple, samba, mamba2hybrid, jamba, yang-2024-parallelizing}, or by distilling pretrained Transformers into hybrid GSA-attention models, as in \citet{wang-2024-mamballama}, or using different training objectives with JRT prompts, as in \citet{arora-2024-jrt}, or combining with YOCO \cite{sun-2024-yoco,Goldstein2024GoldFinchHP}.

GSA follows GLA in using a gated update rule, although we acknowledge recent work on Parallel DeltaNet \cite{yang-2024-parallelizing}, which parallelizes the delta update rule computations in DeltaNet \cite{schlag-2021-deltanet} over sequence length, significantly enhancing training efficiency. The delta rule is known to improve in-context retrieval ability \cite{schlag-2021-deltanet,yang-2024-parallelizing}, aligning with one of the objectives of this work.
We did not explore the analogous two-pass DeltaNet, but we leave this for future investigation, which would bring the approach closer to the original DeltaMLP \cite{Irie-2021-going}, as discussed earlier.
It would also be beneficial to compare GSA with more recent strong RNN models, such as xLSTM \cite{beck-2024-xlstm}, Mamba2 \cite{mamba2}, TTT \cite{sun-2024-learning}, and Longhorn \cite{liu-2024-longhorn}.

\section{Conclusions}

This work introduces Gated Slot Attention (\textsc{GSA}), which enhances ABC \cite{peng-etal-2022-abc} with a gating mechanism inspired by Gated Linear Attention (GLA \cite{yang-etal-2024-gla}). By framing \textsc{GSA} as a two-pass GLA, we can leverage hardware-efficient implementations of GLA \citep{yang-2024-fla} to train GSA. 
As such, \textsc{GSA} benefits from context-aware memory reading and forgetting, implicitly increasing the model's capacity despite a small actual state size, which improves training and inference efficiency. Through extensive experiments, we demonstrate the advantages of \textsc{GSA} in in-context recall-intensive tasks \cite{arora-2024-jrt} and in ``finetuning pretrained Transformers to RNNs'' \cite{kasai-etal-2021-finetuning} scenarios.

\begin{ack}
  We would like to thank Zhen Qin and Yikang Shen for their insightful discussions, Houquan Zhou and Kazuki Irie for providing valuable feedback on this manuscript.

  We gratefully acknowledge the support by LuxiTech for computational resources; and the
  support by the National Natural Science Foundation of China (No. 62076173, 62476187), the High-level Entrepreneurship and Innovation Plan of Jiangsu Province (No. JSSCRC2021524), and the Project Funded by the Priority Academic Program Development of Jiangsu Higher Education Institutions.
  Yu Zhang was partially supported by Tencent AI Lab under Wei Bi's mentorship.
  Songlin Yang was supported by Xianhong Wu Fellowship from MIT.
\end{ack}

\medskip

\bibliography{main}

\begin{thebibliography}{106}
\providecommand{\natexlab}[1]{#1}
\providecommand{\url}[1]{\texttt{#1}}
\expandafter\ifx\csname urlstyle\endcsname\relax
  \providecommand{\doi}[1]{doi: #1}\else
  \providecommand{\doi}{doi: \begingroup \urlstyle{rm}\Url}\fi

\bibitem[Aky{\"{u}}rek et~al.(2024)Aky{\"{u}}rek, Wang, Kim, and Andreas]{akyurek-etal-2024-inctx}
E.~Aky{\"{u}}rek, B.~Wang, Y.~Kim, and J.~Andreas.
\newblock In-context language learning: Architectures and algorithms.
\newblock In \emph{Proceedings of ICML}, 2024.
\newblock URL \url{https://openreview.net/forum?id=3Z9CRr5srL}.

\bibitem[Arora et~al.(2023)Arora, Eyuboglu, Timalsina, Johnson, Poli, Zou, Rudra, and Ré]{arora-2023-zoology}
S.~Arora, S.~Eyuboglu, A.~Timalsina, I.~Johnson, M.~Poli, J.~Zou, A.~Rudra, and C.~Ré.
\newblock Zoology: Measuring and improving recall in efficient language models, 2023.

\bibitem[Arora et~al.(2024{\natexlab{a}})Arora, Eyuboglu, Zhang, Timalsina, Alberti, Zinsley, Zou, Rudra, and Ré]{arora-2024-simple}
S.~Arora, S.~Eyuboglu, M.~Zhang, A.~Timalsina, S.~Alberti, D.~Zinsley, J.~Zou, A.~Rudra, and C.~Ré.
\newblock Simple linear attention language models balance the recall-throughput tradeoff, 2024{\natexlab{a}}.

\bibitem[Arora et~al.(2024{\natexlab{b}})Arora, Timalsina, Singhal, Spector, Eyuboglu, Zhao, Rao, Rudra, and Ré]{arora-2024-jrt}
S.~Arora, A.~Timalsina, A.~Singhal, B.~Spector, S.~Eyuboglu, X.~Zhao, A.~Rao, A.~Rudra, and C.~Ré.
\newblock Just read twice: closing the recall gap for recurrent language models, 2024{\natexlab{b}}.
\newblock URL \url{https://arxiv.org/abs/2407.05483}.

\bibitem[Ba et~al.(2016)Ba, Hinton, Mnih, Leibo, and Ionescu]{ba-2016-using}
J.~Ba, G.~Hinton, V.~Mnih, J.~Z. Leibo, and C.~Ionescu.
\newblock Using fast weights to attend to the recent past, 2016.

\bibitem[Beck et~al.(2024)Beck, Pöppel, Spanring, Auer, Prudnikova, Kopp, Klambauer, Brandstetter, and Hochreiter]{beck-2024-xlstm}
M.~Beck, K.~Pöppel, M.~Spanring, A.~Auer, O.~Prudnikova, M.~Kopp, G.~Klambauer, J.~Brandstetter, and S.~Hochreiter.
\newblock xlstm: Extended long short-term memory, 2024.

\bibitem[Bick et~al.(2024)Bick, Li, Xing, Kolter, and Gu]{bick-2024-transformertossms}
A.~Bick, K.~Y. Li, E.~P. Xing, J.~Z. Kolter, and A.~Gu.
\newblock Transformers to ssms: Distilling quadratic knowledge to subquadratic models.
\newblock 2024.
\newblock URL \url{https://api.semanticscholar.org/CorpusID:271903923}.

\bibitem[Bisk et~al.(2020)Bisk, Zellers, Bras, Gao, and Choi]{bisk-2020-piqa}
Y.~Bisk, R.~Zellers, R.~L. Bras, J.~Gao, and Y.~Choi.
\newblock Piqa: Reasoning about physical commonsense in natural language.
\newblock In \emph{In Proceedings of AAAI}, 2020.

\bibitem[Blelloch(1993)]{blelloch-1993-prefix}
G.~E. Blelloch.
\newblock Prefix sums and their applications.
\newblock Technical report, School of Computer Science, Carnegie Mellon University, Pittsburgh, PA, 1993.
\newblock URL \url{https://www.cs.cmu.edu/~guyb/papers/Ble93.pdf}.

\bibitem[Chen et~al.(2024)Chen, Liu, Wang, Tian, and Wang]{chen-2024-dijiang}
H.~Chen, Z.~Liu, X.~Wang, Y.~Tian, and Y.~Wang.
\newblock Dijiang: Efficient large language models through compact kernelization.
\newblock \emph{ArXiv}, abs/2403.19928, 2024.
\newblock URL \url{https://api.semanticscholar.org/CorpusID:268793982}.

\bibitem[Chen et~al.(2016)Chen, Xu, Zhang, and Guestrin]{chen-2016-training}
T.~Chen, B.~Xu, C.~Zhang, and C.~Guestrin.
\newblock Training deep nets with sublinear memory cost, 2016.

\bibitem[Cho et~al.(2014)Cho, van Merri{\"e}nboer, Gulcehre, Bahdanau, Bougares, Schwenk, and Bengio]{cho-etal-2014-gru}
K.~Cho, B.~van Merri{\"e}nboer, C.~Gulcehre, D.~Bahdanau, F.~Bougares, H.~Schwenk, and Y.~Bengio.
\newblock Learning phrase representations using {RNN} encoder{--}decoder for statistical machine translation.
\newblock In \emph{Proceedings of EMNLP}, pages 1724--1734, 2014.
\newblock URL \url{https://aclanthology.org/D14-1179}.

\bibitem[Choi(2024)]{choi-2024-cross}
S.~Choi.
\newblock Cross-architecture transfer learning for linear-cost inference transformers, 2024.
\newblock URL \url{https://arxiv.org/abs/2404.02684}.

\bibitem[Clark et~al.(2018)Clark, Cowhey, Etzioni, Khot, Sabharwal, Schoenick, and Tafjord]{clark-2018-arc}
P.~Clark, I.~Cowhey, O.~Etzioni, T.~Khot, A.~Sabharwal, C.~Schoenick, and O.~Tafjord.
\newblock Think you have solved question answering? try arc, the ai2 reasoning challenge.
\newblock \emph{arXiv:1803.05457v1}, 2018.

\bibitem[Dao(2024)]{dao-2024-flashattn2}
T.~Dao.
\newblock Flashattention-2: Faster attention with better parallelism and work partitioning.
\newblock In \emph{The Twelfth International Conference on Learning Representations}, 2024.
\newblock URL \url{https://openreview.net/forum?id=mZn2Xyh9Ec}.

\bibitem[Dao and Gu(2024)]{mamba2}
T.~Dao and A.~Gu.
\newblock Transformers are ssms: Generalized models and efficient algorithms through structured state space duality.
\newblock \emph{CoRR}, abs/2405.21060, 2024.
\newblock \doi{10.48550/ARXIV.2405.21060}.
\newblock URL \url{https://doi.org/10.48550/arXiv.2405.21060}.

\bibitem[Dao et~al.(2022)Dao, Fu, Ermon, Rudra, and R\'{e}]{dao-2022-flashattn}
T.~Dao, D.~Fu, S.~Ermon, A.~Rudra, and C.~R\'{e}.
\newblock Flashattention: Fast and memory-efficient exact attention with io-awareness.
\newblock In \emph{Advances in NIPS}, pages 16344--16359, 2022.
\newblock URL \url{https://proceedings.neurips.cc/paper_files/paper/2022/file/67d57c32e20fd0a7a302cb81d36e40d5-Paper-Conference.pdf}.

\bibitem[Dasigi et~al.(2021)Dasigi, Lo, Beltagy, Cohan, Smith, and Gardner]{dasigi-etal-2021-dataset}
P.~Dasigi, K.~Lo, I.~Beltagy, A.~Cohan, N.~A. Smith, and M.~Gardner.
\newblock A dataset of information-seeking questions and answers anchored in research papers.
\newblock In \emph{Proceedings of NAACL}, pages 4599--4610, 2021.
\newblock URL \url{https://aclanthology.org/2021.naacl-main.365}.

\bibitem[Dauphin et~al.(2017)Dauphin, Fan, Auli, and Grangier]{dauphin-2017-glu}
Y.~N. Dauphin, A.~Fan, M.~Auli, and D.~Grangier.
\newblock Language modeling with gated convolutional networks.
\newblock In \emph{Proceedings of ICML}, pages 933--941, 2017.
\newblock URL \url{https://proceedings.mlr.press/v70/dauphin17a.html}.

\bibitem[Dua et~al.(2019)Dua, Wang, Dasigi, Stanovsky, Singh, and Gardner]{dua-etal-2019-drop}
D.~Dua, Y.~Wang, P.~Dasigi, G.~Stanovsky, S.~Singh, and M.~Gardner.
\newblock {DROP}: A reading comprehension benchmark requiring discrete reasoning over paragraphs.
\newblock In \emph{Proceedings of NAACL}, pages 2368--2378, 2019.
\newblock URL \url{https://aclanthology.org/N19-1246}.

\bibitem[Gao et~al.(2023)Gao, Tow, Abbasi, Biderman, Black, DiPofi, Foster, Golding, Hsu, Le~Noac'h, Li, McDonell, Muennighoff, Ociepa, Phang, Reynolds, Schoelkopf, Skowron, Sutawika, Tang, Thite, Wang, Wang, and Zou]{gao-2023-eval-harness}
L.~Gao, J.~Tow, B.~Abbasi, S.~Biderman, S.~Black, A.~DiPofi, C.~Foster, L.~Golding, J.~Hsu, A.~Le~Noac'h, H.~Li, K.~McDonell, N.~Muennighoff, C.~Ociepa, J.~Phang, L.~Reynolds, H.~Schoelkopf, A.~Skowron, L.~Sutawika, E.~Tang, A.~Thite, B.~Wang, K.~Wang, and A.~Zou.
\newblock A framework for few-shot language model evaluation, 2023.
\newblock URL \url{https://zenodo.org/records/10256836}.

\bibitem[Gers et~al.(1999)Gers, Schmidhuber, and Cummins]{gers-1999-forget}
F.~Gers, J.~Schmidhuber, and F.~Cummins.
\newblock Learning to forget: continual prediction with lstm.
\newblock In \emph{Proceedings of ICANN}, pages 850--855, 1999.

\bibitem[Gerstenberger et~al.(2020)Gerstenberger, Irie, Golik, Beck, and Ney]{gerstenberger-2020-icassp-domain}
A.~Gerstenberger, K.~Irie, P.~Golik, E.~Beck, and H.~Ney.
\newblock Domain robust, fast, and compact neural language models.
\newblock In \emph{Proceedings of ICASSP}, Barcelona, Spain, 2020.

\bibitem[Geva et~al.(2021)Geva, Schuster, Berant, and Levy]{geva-etal-2021-transformer}
M.~Geva, R.~Schuster, J.~Berant, and O.~Levy.
\newblock Transformer feed-forward layers are key-value memories.
\newblock In \emph{Proceedings of EMNLP}, pages 5484--5495, Online and Punta Cana, Dominican Republic, 2021.
\newblock URL \url{https://aclanthology.org/2021.emnlp-main.446}.

\bibitem[Goldstein et~al.(2024)Goldstein, Obeid, Alcaide, Song, and Cheah]{Goldstein2024GoldFinchHP}
D.~Goldstein, F.~Obeid, E.~Alcaide, G.~Song, and E.~Cheah.
\newblock Goldfinch: High performance rwkv/transformer hybrid with linear pre-fill and extreme kv-cache compression.
\newblock \emph{ArXiv}, abs/2407.12077, 2024.
\newblock URL \url{https://api.semanticscholar.org/CorpusID:271244694}.

\bibitem[Graves(2014)]{graves-2014-generating}
A.~Graves.
\newblock Generating sequences with recurrent neural networks, 2014.

\bibitem[Graves et~al.(2014)Graves, Wayne, and Danihelka]{graves-2014-neural}
A.~Graves, G.~Wayne, and I.~Danihelka.
\newblock Neural turing machines, 2014.

\bibitem[Grazzi et~al.(2024)Grazzi, Siems, Schrodi, Brox, and Hutter]{Grazzi2024IsMC}
R.~Grazzi, J.~N. Siems, S.~Schrodi, T.~Brox, and F.~Hutter.
\newblock Is mamba capable of in-context learning?
\newblock \emph{ArXiv}, abs/2402.03170, 2024.
\newblock URL \url{https://api.semanticscholar.org/CorpusID:267412719}.

\bibitem[Gu and Dao(2023)]{gu-2023-mamba}
A.~Gu and T.~Dao.
\newblock Mamba: Linear-time sequence modeling with selective state spaces, 2023.

\bibitem[Gu et~al.(2020)Gu, Gulcehre, Paine, Hoffman, and Pascanu]{gu-2020-gating}
A.~Gu, C.~Gulcehre, T.~Paine, M.~Hoffman, and R.~Pascanu.
\newblock Improving the gating mechanism of recurrent neural networks.
\newblock In H.~D. III and A.~Singh, editors, \emph{Proceedings of ICML}, pages 3800--3809. PMLR, 2020.
\newblock URL \url{https://proceedings.mlr.press/v119/gu20a.html}.

\bibitem[Hendrycks et~al.(2021)Hendrycks, Burns, Basart, Zou, Mazeika, Song, and Steinhardt]{hendrycks-2021-mmlu}
D.~Hendrycks, C.~Burns, S.~Basart, A.~Zou, M.~Mazeika, D.~Song, and J.~Steinhardt.
\newblock Measuring massive multitask language understanding, 2021.
\newblock URL \url{https://arxiv.org/abs/2009.03300}.

\bibitem[Hochreiter and Schmidhuber(1997)]{hochreiter-1997-lstm}
S.~Hochreiter and J.~Schmidhuber.
\newblock Long short-term memory.
\newblock \emph{Neural Computation}, 9\penalty0 (8):\penalty0 1735--1780, 1997.

\bibitem[Hua et~al.(2022)Hua, Dai, Liu, and Le]{hua-etal-2022-gau}
W.~Hua, Z.~Dai, H.~Liu, and Q.~Le.
\newblock Transformer quality in linear time.
\newblock In K.~Chaudhuri, S.~Jegelka, L.~Song, C.~Szepesvari, G.~Niu, and S.~Sabato, editors, \emph{Proceedings of ICML}, pages 9099--9117. PMLR, 2022.
\newblock URL \url{https://proceedings.mlr.press/v162/hua22a.html}.

\bibitem[Irie et~al.(2020)Irie, Gerstenberger, Schlüter, and Ney]{irie-2020-how}
K.~Irie, A.~Gerstenberger, R.~Schlüter, and H.~Ney.
\newblock How much self-attention do we need? {T}rading attention for feed-forward layers.
\newblock In \emph{Proceedings of ICASSP}, Virtual only, May 2020.

\bibitem[Irie et~al.(2021)Irie, Schlag, Csord'as, and Schmidhuber]{Irie-2021-going}
K.~Irie, I.~Schlag, R.~Csord'as, and J.~Schmidhuber.
\newblock Going beyond linear transformers with recurrent fast weight programmers.
\newblock \emph{ArXiv}, abs/2106.06295, 2021.
\newblock URL \url{https://api.semanticscholar.org/CorpusID:235417174}.

\bibitem[Irie et~al.(2022)Irie, Schlag, Csord'as, and Schmidhuber]{Irie-2022-ams}
K.~Irie, I.~Schlag, R.~Csord'as, and J.~Schmidhuber.
\newblock A modern self-referential weight matrix that learns to modify itself.
\newblock In \emph{International Conference on Machine Learning}, 2022.
\newblock URL \url{https://api.semanticscholar.org/CorpusID:246823084}.

\bibitem[Jelassi et~al.(2024{\natexlab{a}})Jelassi, Brandfonbrener, Kakade, and Malach]{DBLP:conf/icml/JelassiBKM24}
S.~Jelassi, D.~Brandfonbrener, S.~M. Kakade, and E.~Malach.
\newblock Repeat after me: Transformers are better than state space models at copying.
\newblock In \emph{Forty-first International Conference on Machine Learning, {ICML} 2024, Vienna, Austria, July 21-27, 2024}. OpenReview.net, 2024{\natexlab{a}}.
\newblock URL \url{https://openreview.net/forum?id=duRRoGeoQT}.

\bibitem[Jelassi et~al.(2024{\natexlab{b}})Jelassi, Brandfonbrener, Kakade, and Malach]{jelassi-2024-repeat}
S.~Jelassi, D.~Brandfonbrener, S.~M. Kakade, and E.~Malach.
\newblock Repeat after me: Transformers are better than state space models at copying, 2024{\natexlab{b}}.

\bibitem[Jiang et~al.(2023)Jiang, Sablayrolles, Mensch, Bamford, Chaplot, de~las Casas, Bressand, Lengyel, Lample, Saulnier, Lavaud, Lachaux, Stock, Scao, Lavril, Wang, Lacroix, and Sayed]{jiang-2023-mistral}
A.~Q. Jiang, A.~Sablayrolles, A.~Mensch, C.~Bamford, D.~S. Chaplot, D.~de~las Casas, F.~Bressand, G.~Lengyel, G.~Lample, L.~Saulnier, L.~R. Lavaud, M.-A. Lachaux, P.~Stock, T.~L. Scao, T.~Lavril, T.~Wang, T.~Lacroix, and W.~E. Sayed.
\newblock Mistral 7b, 2023.

\bibitem[Joshi et~al.(2017)Joshi, Choi, Weld, and Zettlemoyer]{joshi-etal-2017-triviaqa}
M.~Joshi, E.~Choi, D.~Weld, and L.~Zettlemoyer.
\newblock {T}rivia{QA}: A large scale distantly supervised challenge dataset for reading comprehension.
\newblock In R.~Barzilay and M.-Y. Kan, editors, \emph{Proceedings of ACL}, pages 1601--1611, Vancouver, Canada, July 2017.
\newblock \doi{10.18653/v1/P17-1147}.
\newblock URL \url{https://aclanthology.org/P17-1147}.

\bibitem[Kasai et~al.(2021)Kasai, Peng, Zhang, Yogatama, Ilharco, Pappas, Mao, Chen, and Smith]{kasai-etal-2021-finetuning}
J.~Kasai, H.~Peng, Y.~Zhang, D.~Yogatama, G.~Ilharco, N.~Pappas, Y.~Mao, W.~Chen, and N.~A. Smith.
\newblock Finetuning pretrained transformers into {RNN}s.
\newblock In M.-F. Moens, X.~Huang, L.~Specia, and S.~W.-t. Yih, editors, \emph{Proceedings of EMNLP}, pages 10630--10643, 2021.
\newblock URL \url{https://aclanthology.org/2021.emnlp-main.830}.

\bibitem[Katharopoulos et~al.(2020)Katharopoulos, Vyas, Pappas, and Fleuret]{katharopoulos-2020-transformers}
A.~Katharopoulos, A.~Vyas, N.~Pappas, and F.~Fleuret.
\newblock Transformers are {RNN}s: Fast autoregressive transformers with linear attention.
\newblock In H.~D. III and A.~Singh, editors, \emph{Proceedings of ICML}, pages 5156--5165. PMLR, 2020.
\newblock URL \url{https://proceedings.mlr.press/v119/katharopoulos20a.html}.

\bibitem[Katsch(2024)]{katsch-2024-gateloop}
T.~Katsch.
\newblock Gateloop: Fully data-controlled linear recurrence for sequence modeling, 2024.

\bibitem[Ko{\v{c}}isk{\'y} et~al.(2018)Ko{\v{c}}isk{\'y}, Schwarz, Blunsom, Dyer, Hermann, Melis, and Grefenstette]{kocisky-etal-2018-narrativeqa}
T.~Ko{\v{c}}isk{\'y}, J.~Schwarz, P.~Blunsom, C.~Dyer, K.~M. Hermann, G.~Melis, and E.~Grefenstette.
\newblock The {N}arrative{QA} reading comprehension challenge.
\newblock \emph{TACL}, pages 317--328, 2018.
\newblock URL \url{https://aclanthology.org/Q18-1023}.

\bibitem[Krotov and Hopfield(2021)]{krotov_large_2021}
D.~Krotov and J.~Hopfield.
\newblock Large {Associative} {Memory} {Problem} in {Neurobiology} and {Machine} {Learning}, 2021.
\newblock URL \url{http://arxiv.org/abs/2008.06996}.

\bibitem[Kwiatkowski et~al.(2019)Kwiatkowski, Palomaki, Redfield, Collins, Parikh, Alberti, Epstein, Polosukhin, Devlin, Lee, Toutanova, Jones, Kelcey, Chang, Dai, Uszkoreit, Le, and Petrov]{kwiatkowski-etal-2019-natural}
T.~Kwiatkowski, J.~Palomaki, O.~Redfield, M.~Collins, A.~Parikh, C.~Alberti, D.~Epstein, I.~Polosukhin, J.~Devlin, K.~Lee, K.~Toutanova, L.~Jones, M.~Kelcey, M.-W. Chang, A.~M. Dai, J.~Uszkoreit, Q.~Le, and S.~Petrov.
\newblock Natural questions: A benchmark for question answering research.
\newblock \emph{TACL}, pages 452--466, 2019.
\newblock URL \url{https://aclanthology.org/Q19-1026}.

\bibitem[Lieber et~al.(2024)Lieber, Lenz, Bata, Cohen, Osin, Dalmedigos, Safahi, Meirom, Belinkov, Shalev-Shwartz, Abend, Alon, Asida, Bergman, Glozman, Gokhman, Manevich, Ratner, Rozen, Shwartz, Zusman, and Shoham]{jamba}
O.~Lieber, B.~Lenz, H.~Bata, G.~Cohen, J.~Osin, I.~Dalmedigos, E.~Safahi, S.~Meirom, Y.~Belinkov, S.~Shalev-Shwartz, O.~Abend, R.~Alon, T.~Asida, A.~Bergman, R.~Glozman, M.~Gokhman, A.~Manevich, N.~Ratner, N.~Rozen, E.~Shwartz, M.~Zusman, and Y.~Shoham.
\newblock Jamba: A hybrid transformer-mamba language model, 2024.
\newblock URL \url{https://arxiv.org/abs/2403.19887}.

\bibitem[Liu et~al.(2024)Liu, Wang, Wu, Feng, Stone, and Liu]{liu-2024-longhorn}
B.~Liu, R.~Wang, L.~Wu, Y.~Feng, P.~Stone, and Q.~Liu.
\newblock Longhorn: State space models are amortized online learners.
\newblock \emph{ArXiv}, abs/2407.14207, 2024.
\newblock URL \url{https://api.semanticscholar.org/CorpusID:271310065}.

\bibitem[Lockard et~al.(2019)Lockard, Shiralkar, and Dong]{lockard-2019-openceres}
C.~Lockard, P.~Shiralkar, and X.~L. Dong.
\newblock {OpenCeres}: {When} {Open} {Information} {Extraction} {Meets} the {Semi}-{Structured} {Web}.
\newblock In J.~Burstein, C.~Doran, and T.~Solorio, editors, \emph{Proceedings of NAACL}, pages 3047--3056, Minneapolis, Minnesota, 2019.
\newblock \doi{10.18653/v1/N19-1309}.
\newblock URL \url{https://aclanthology.org/N19-1309}.

\bibitem[Loshchilov and Hutter(2019)]{loshchilov-2019-decoupled}
I.~Loshchilov and F.~Hutter.
\newblock Decoupled weight decay regularization, 2019.

\bibitem[Ma et~al.(2021)Ma, Kong, Wang, Zhou, May, Ma, and Zettlemoyer]{ma-2021-luna}
X.~Ma, X.~Kong, S.~Wang, C.~Zhou, J.~May, H.~Ma, and L.~Zettlemoyer.
\newblock Luna: Linear unified nested attention.
\newblock In M.~Ranzato, A.~Beygelzimer, Y.~Dauphin, P.~Liang, and J.~W. Vaughan, editors, \emph{Advances in NIPS}, volume~34, pages 2441--2453. Curran Associates, Inc., 2021.
\newblock URL \url{https://proceedings.neurips.cc/paper_files/paper/2021/file/14319d9cfc6123106878dc20b94fbaf3-Paper.pdf}.

\bibitem[Mao(2022)]{mao-2022-fine}
H.~H. Mao.
\newblock Fine-tuning pre-trained transformers into decaying fast weights.
\newblock In Y.~Goldberg, Z.~Kozareva, and Y.~Zhang, editors, \emph{Proceedings of EMNLP}, pages 10236--10242, Abu Dhabi, United Arab Emirates, 2022.
\newblock URL \url{https://aclanthology.org/2022.emnlp-main.697}.

\bibitem[Martin and Cundy(2018)]{martin-2018-parallelizing}
E.~Martin and C.~Cundy.
\newblock Parallelizing linear recurrent neural nets over sequence length.
\newblock In \emph{Proceedings of ICLR}, 2018.
\newblock URL \url{https://openreview.net/forum?id=HyUNwulC-}.

\bibitem[Mercat et~al.(2024)Mercat, Vasiljevic, Keh, Arora, Dave, Gaidon, and Kollar]{mercat-2024-linearizing}
J.~Mercat, I.~Vasiljevic, S.~Keh, K.~Arora, A.~Dave, A.~Gaidon, and T.~Kollar.
\newblock Linearizing large language models, 2024.

\bibitem[Merity et~al.(2016)Merity, Xiong, Bradbury, and Socher]{merity-2016-pointer}
S.~Merity, C.~Xiong, J.~Bradbury, and R.~Socher.
\newblock Pointer sentinel mixture models, 2016.

\bibitem[Munkhdalai et~al.(2019)Munkhdalai, Sordoni, Wang, and Trischler]{munkhdalai-2019-metalearned}
T.~Munkhdalai, A.~Sordoni, T.~Wang, and A.~Trischler.
\newblock Metalearned neural memory.
\newblock \emph{ArXiv}, abs/1907.09720, 2019.
\newblock URL \url{https://api.semanticscholar.org/CorpusID:198179407}.

\bibitem[Oren et~al.(2024)Oren, Hassid, Adi, and Schwartz]{oren-2024-transformers}
M.~Oren, M.~Hassid, Y.~Adi, and R.~Schwartz.
\newblock Transformers are multi-state rnns, 2024.

\bibitem[Pang et~al.(2022)Pang, Parrish, Joshi, Nangia, Phang, Chen, Padmakumar, Ma, Thompson, He, and Bowman]{pang-2022-quality}
R.~Y. Pang, A.~Parrish, N.~Joshi, N.~Nangia, J.~Phang, A.~Chen, V.~Padmakumar, J.~Ma, J.~Thompson, H.~He, and S.~R. Bowman.
\newblock Quality: Question answering with long input texts, yes!, 2022.
\newblock URL \url{https://arxiv.org/abs/2112.08608}.

\bibitem[Paperno et~al.(2016)Paperno, Kruszewski, Lazaridou, Pham, Bernardi, Pezzelle, Baroni, Boleda, and Fernández]{paperno-2019-lambada}
D.~Paperno, G.~Kruszewski, A.~Lazaridou, Q.~N. Pham, R.~Bernardi, S.~Pezzelle, M.~Baroni, G.~Boleda, and R.~Fernández.
\newblock The lambada dataset, 2016.

\bibitem[Penedo et~al.(2023)Penedo, Malartic, Hesslow, Cojocaru, Cappelli, Alobeidli, Pannier, Almazrouei, and Launay]{zhang-2023-refinedweb}
G.~Penedo, Q.~Malartic, D.~Hesslow, R.~Cojocaru, A.~Cappelli, H.~Alobeidli, B.~Pannier, E.~Almazrouei, and J.~Launay.
\newblock The {R}efined{W}eb dataset for {F}alcon {LLM}: outperforming curated corpora with web data, and web data only.
\newblock \emph{arXiv preprint arXiv:2306.01116}, 2023.
\newblock URL \url{https://arxiv.org/abs/2306.01116}.

\bibitem[Peng et~al.(2024)Peng, Goldstein, Anthony, Albalak, Alcaide, Biderman, Cheah, Du, Ferdinan, Hou, Kazienko, GV, Kocoń, Koptyra, Krishna, au2, Muennighoff, Obeid, Saito, Song, Tu, Woźniak, Zhang, Zhao, Zhao, Zhou, Zhu, and Zhu]{peng-2024-eagle}
B.~Peng, D.~Goldstein, Q.~Anthony, A.~Albalak, E.~Alcaide, S.~Biderman, E.~Cheah, X.~Du, T.~Ferdinan, H.~Hou, P.~Kazienko, K.~K. GV, J.~Kocoń, B.~Koptyra, S.~Krishna, R.~M.~J. au2, N.~Muennighoff, F.~Obeid, A.~Saito, G.~Song, H.~Tu, S.~Woźniak, R.~Zhang, B.~Zhao, Q.~Zhao, P.~Zhou, J.~Zhu, and R.-J. Zhu.
\newblock Eagle and finch: Rwkv with matrix-valued states and dynamic recurrence, 2024.

\bibitem[Peng et~al.(2021)Peng, Pappas, Yogatama, Schwartz, Smith, and Kong]{peng-2021-rfa}
H.~Peng, N.~Pappas, D.~Yogatama, R.~Schwartz, N.~Smith, and L.~Kong.
\newblock Random feature attention.
\newblock In \emph{Proceedings of ICLR}, 2021.
\newblock URL \url{https://openreview.net/forum?id=QtTKTdVrFBB}.

\bibitem[Peng et~al.(2022)Peng, Kasai, Pappas, Yogatama, Wu, Kong, Schwartz, and Smith]{peng-etal-2022-abc}
H.~Peng, J.~Kasai, N.~Pappas, D.~Yogatama, Z.~Wu, L.~Kong, R.~Schwartz, and N.~A. Smith.
\newblock {ABC}: Attention with bounded-memory control.
\newblock In \emph{Proceedings of ACL}, pages 7469--7483, 2022.
\newblock URL \url{https://aclanthology.org/2022.acl-long.515}.

\bibitem[Pope et~al.(2022)Pope, Douglas, Chowdhery, Devlin, Bradbury, Levskaya, Heek, Xiao, Agrawal, and Dean]{pope-2022-efficiently}
R.~Pope, S.~Douglas, A.~Chowdhery, J.~Devlin, J.~Bradbury, A.~Levskaya, J.~Heek, K.~Xiao, S.~Agrawal, and J.~Dean.
\newblock Efficiently scaling transformer inference, 2022.

\bibitem[Pramanik et~al.(2023)Pramanik, Elelimy, Machado, and White]{pramanik-2023-recurrentlineartransformers}
S.~Pramanik, E.~Elelimy, M.~C. Machado, and A.~White.
\newblock Recurrent linear transformers, 2023.
\newblock URL \url{https://arxiv.org/abs/2310.15719}.

\bibitem[Qin et~al.(2022)Qin, Han, Sun, Li, Kong, Barnes, and Zhong]{qin-etal-2022-devil}
Z.~Qin, X.~Han, W.~Sun, D.~Li, L.~Kong, N.~Barnes, and Y.~Zhong.
\newblock The devil in linear transformer.
\newblock In Y.~Goldberg, Z.~Kozareva, and Y.~Zhang, editors, \emph{Proceedings of EMNLP}, pages 7025--7041, Abu Dhabi, United Arab Emirates, 2022.
\newblock \doi{10.18653/v1/2022.emnlp-main.473}.
\newblock URL \url{https://aclanthology.org/2022.emnlp-main.473}.

\bibitem[Qin et~al.(2023)Qin, Yang, and Zhong]{qin-2023-hgrn}
Z.~Qin, S.~Yang, and Y.~Zhong.
\newblock Hierarchically gated recurrent neural network for sequence modeling.
\newblock In \emph{Advances in NIPS}, 2023.
\newblock URL \url{https://openreview.net/forum?id=P1TCHxJwLB}.

\bibitem[Qin et~al.(2024{\natexlab{a}})Qin, Li, Sun, Sun, Shen, Han, Wei, Lv, Luo, Qiao, and Zhong]{qin-2024-transnormerllm}
Z.~Qin, D.~Li, W.~Sun, W.~Sun, X.~Shen, X.~Han, Y.~Wei, B.~Lv, X.~Luo, Y.~Qiao, and Y.~Zhong.
\newblock Transnormerllm: A faster and better large language model with improved transnormer, 2024{\natexlab{a}}.

\bibitem[Qin et~al.(2024{\natexlab{b}})Qin, Yang, Sun, Shen, Li, Sun, and Zhong]{qin-2024-hgrn2}
Z.~Qin, S.~Yang, W.~Sun, X.~Shen, D.~Li, W.~Sun, and Y.~Zhong.
\newblock Hgrn2: Gated linear rnns with state expansion, 2024{\natexlab{b}}.

\bibitem[Rajpurkar et~al.(2018)Rajpurkar, Jia, and Liang]{rajpurkar-2018-know}
P.~Rajpurkar, R.~Jia, and P.~Liang.
\newblock Know {What} {You} {Don}'t {Know}: {Unanswerable} {Questions} for {SQuAD}.
\newblock In \emph{Proceedings of ACL}, Melbourne, Australia, 2018. Association for Computational Linguistics.

\bibitem[Ren et~al.(2024)Ren, Liu, Lu, Shen, Liang, and Chen]{samba}
L.~Ren, Y.~Liu, Y.~Lu, Y.~Shen, C.~Liang, and W.~Chen.
\newblock Samba: Simple hybrid state space models for efficient unlimited context language modeling.
\newblock \emph{CoRR}, abs/2406.07522, 2024.
\newblock \doi{10.48550/ARXIV.2406.07522}.
\newblock URL \url{https://doi.org/10.48550/arXiv.2406.07522}.

\bibitem[Rush(2020)]{rush-2020-torch}
A.~Rush.
\newblock Torch-struct: Deep structured prediction library.
\newblock In A.~Celikyilmaz and T.-H. Wen, editors, \emph{Proceedings of ACL}, pages 335--342, Online, July 2020.
\newblock \doi{10.18653/v1/2020.acl-demos.38}.
\newblock URL \url{https://aclanthology.org/2020.acl-demos.38}.

\bibitem[Sakaguchi et~al.(2019)Sakaguchi, Bras, Bhagavatula, and Choi]{keisuke-2019-winogrande}
K.~Sakaguchi, R.~L. Bras, C.~Bhagavatula, and Y.~Choi.
\newblock Winogrande: An adversarial winograd schema challenge at scale, 2019.
\newblock URL \url{https://arxiv.org/abs/1907.10641}.

\bibitem[Schlag and Schmidhuber(2017)]{schlag-2017-gated}
I.~Schlag and J.~Schmidhuber.
\newblock Gated fast weights for on-the-fly neural program generation.
\newblock In \emph{Proceedings of ICLR}, 2017.
\newblock URL \url{https://api.semanticscholar.org/CorpusID:216094255}.

\bibitem[Schlag et~al.(2021{\natexlab{a}})Schlag, Irie, and Schmidhuber]{schlag-2021-deltanet}
I.~Schlag, K.~Irie, and J.~Schmidhuber.
\newblock Linear transformers are secretly fast weight programmers.
\newblock In M.~Meila and T.~Zhang, editors, \emph{Proceedings of ICML}, pages 9355--9366. PMLR, 18--24 Jul 2021{\natexlab{a}}.
\newblock URL \url{https://proceedings.mlr.press/v139/schlag21a.html}.

\bibitem[Schlag et~al.(2021{\natexlab{b}})Schlag, Munkhdalai, and Schmidhuber]{schlag-2021-learning}
I.~Schlag, T.~Munkhdalai, and J.~Schmidhuber.
\newblock Learning associative inference using fast weight memory, 2021{\natexlab{b}}.
\newblock URL \url{https://arxiv.org/abs/2011.07831}.

\bibitem[Schmidhuber(1992)]{schmidhuber1992learning}
J.~Schmidhuber.
\newblock Learning to control fast-weight memories: An alternative to dynamic recurrent networks.
\newblock \emph{Neural Computation}, 4\penalty0 (1):\penalty0 131--139, 1992.

\bibitem[Shazeer(2020)]{shazeer-2020-glu}
N.~Shazeer.
\newblock Glu variants improve transformer, 2020.

\bibitem[Soboleva et~al.(2023)Soboleva, Al-Khateeb, Myers, Steeves, Hestness, and Dey]{cerebras-2023-slimpajama}
D.~Soboleva, F.~Al-Khateeb, R.~Myers, J.~R. Steeves, J.~Hestness, and N.~Dey.
\newblock Slimpajama: A 627b token cleaned and deduplicated version of redpajama, 2023.
\newblock URL \url{https://huggingface.co/datasets/cerebras/SlimPajama-627B}.

\bibitem[Su et~al.(2023)Su, Lu, Pan, Murtadha, Wen, and Liu]{su-2023-roformer}
J.~Su, Y.~Lu, S.~Pan, A.~Murtadha, B.~Wen, and Y.~Liu.
\newblock Roformer: Enhanced transformer with rotary position embedding, 2023.

\bibitem[Sukhbaatar et~al.(2015)Sukhbaatar, szlam, Weston, and Fergus]{sukhbaatar-2015-memory}
S.~Sukhbaatar, a.~szlam, J.~Weston, and R.~Fergus.
\newblock End-to-end memory networks.
\newblock In C.~Cortes, N.~Lawrence, D.~Lee, M.~Sugiyama, and R.~Garnett, editors, \emph{Advances in NIPS}. Curran Associates, Inc., 2015.
\newblock URL \url{https://proceedings.neurips.cc/paper_files/paper/2015/file/8fb21ee7a2207526da55a679f0332de2-Paper.pdf}.

\bibitem[Sun et~al.(2023)Sun, Dong, Huang, Ma, Xia, Xue, Wang, and Wei]{sun-2023-retnet}
Y.~Sun, L.~Dong, S.~Huang, S.~Ma, Y.~Xia, J.~Xue, J.~Wang, and F.~Wei.
\newblock Retentive network: A successor to transformer for large language models, 2023.

\bibitem[Sun et~al.(2024{\natexlab{a}})Sun, Dong, Zhu, Huang, Wang, Ma, Zhang, Wang, and Wei]{sun-2024-yoco}
Y.~Sun, L.~Dong, Y.~Zhu, S.~Huang, W.~Wang, S.~Ma, Q.~Zhang, J.~Wang, and F.~Wei.
\newblock You only cache once: Decoder-decoder architectures for language models, 2024{\natexlab{a}}.
\newblock URL \url{https://arxiv.org/abs/2405.05254}.

\bibitem[Sun et~al.(2024{\natexlab{b}})Sun, Li, Dalal, Xu, Vikram, Zhang, Dubois, Chen, Wang, Koyejo, Hashimoto, and Guestrin]{sun-2024-learning}
Y.~Sun, X.~Li, K.~Dalal, J.~Xu, A.~Vikram, G.~Zhang, Y.~Dubois, X.~Chen, X.~Wang, O.~Koyejo, T.~Hashimoto, and C.~Guestrin.
\newblock Learning to (learn at test time): Rnns with expressive hidden states.
\newblock \emph{ArXiv}, abs/2407.04620, 2024{\natexlab{b}}.
\newblock URL \url{https://api.semanticscholar.org/CorpusID:271039606}.

\bibitem[Suzgun et~al.(2023)Suzgun, Scales, Sch{\"a}rli, Gehrmann, Tay, Chung, Chowdhery, Le, Chi, Zhou, and Wei]{suzgun-etal-2023-challenging}
M.~Suzgun, N.~Scales, N.~Sch{\"a}rli, S.~Gehrmann, Y.~Tay, H.~W. Chung, A.~Chowdhery, Q.~Le, E.~Chi, D.~Zhou, and J.~Wei.
\newblock Challenging {BIG}-bench tasks and whether chain-of-thought can solve them.
\newblock In \emph{Findings of the ACL}, pages 13003--13051, Toronto, Canada, 2023.
\newblock URL \url{https://aclanthology.org/2023.findings-acl.824}.

\bibitem[Touvron et~al.(2023{\natexlab{a}})Touvron, Lavril, Izacard, Martinet, Lachaux, Lacroix, Rozière, Goyal, Hambro, Azhar, Rodriguez, Joulin, Grave, and Lample]{touvron-2023-llama}
H.~Touvron, T.~Lavril, G.~Izacard, X.~Martinet, M.-A. Lachaux, T.~Lacroix, B.~Rozière, N.~Goyal, E.~Hambro, F.~Azhar, A.~Rodriguez, A.~Joulin, E.~Grave, and G.~Lample.
\newblock Llama: Open and efficient foundation language models, 2023{\natexlab{a}}.

\bibitem[Touvron et~al.(2023{\natexlab{b}})Touvron, Martin, Stone, Albert, Almahairi, Babaei, Bashlykov, Batra, Bhargava, Bhosale, Bikel, Blecher, Ferrer, Chen, Cucurull, Esiobu, Fernandes, Fu, Fu, Fuller, Gao, Goswami, Goyal, Hartshorn, Hosseini, Hou, Inan, Kardas, Kerkez, Khabsa, Kloumann, Korenev, Koura, Lachaux, Lavril, Lee, Liskovich, Lu, Mao, Martinet, Mihaylov, Mishra, Molybog, Nie, Poulton, Reizenstein, Rungta, Saladi, Schelten, Silva, Smith, Subramanian, Tan, Tang, Taylor, Williams, Kuan, Xu, Yan, Zarov, Zhang, Fan, Kambadur, Narang, Rodriguez, Stojnic, Edunov, and Scialom]{touvron-2023-llama2}
H.~Touvron, L.~Martin, K.~Stone, P.~Albert, A.~Almahairi, Y.~Babaei, N.~Bashlykov, S.~Batra, P.~Bhargava, S.~Bhosale, D.~Bikel, L.~Blecher, C.~C. Ferrer, M.~Chen, G.~Cucurull, D.~Esiobu, J.~Fernandes, J.~Fu, W.~Fu, B.~Fuller, C.~Gao, V.~Goswami, N.~Goyal, A.~Hartshorn, S.~Hosseini, R.~Hou, H.~Inan, M.~Kardas, V.~Kerkez, M.~Khabsa, I.~Kloumann, A.~Korenev, P.~S. Koura, M.-A. Lachaux, T.~Lavril, J.~Lee, D.~Liskovich, Y.~Lu, Y.~Mao, X.~Martinet, T.~Mihaylov, P.~Mishra, I.~Molybog, Y.~Nie, A.~Poulton, J.~Reizenstein, R.~Rungta, K.~Saladi, A.~Schelten, R.~Silva, E.~M. Smith, R.~Subramanian, X.~E. Tan, B.~Tang, R.~Taylor, A.~Williams, J.~X. Kuan, P.~Xu, Z.~Yan, I.~Zarov, Y.~Zhang, A.~Fan, M.~Kambadur, S.~Narang, A.~Rodriguez, R.~Stojnic, S.~Edunov, and T.~Scialom.
\newblock Llama 2: Open foundation and fine-tuned chat models, 2023{\natexlab{b}}.
\newblock URL \url{https://arxiv.org/abs/2307.09288}.

\bibitem[Vaswani et~al.(2017)Vaswani, Shazeer, Parmar, Uszkoreit, Jones, Gomez, Kaiser, and Polosukhin]{vaswani-2017-attention}
A.~Vaswani, N.~Shazeer, N.~Parmar, J.~Uszkoreit, L.~Jones, A.~N. Gomez, L.~u. Kaiser, and I.~Polosukhin.
\newblock Attention is all you need.
\newblock In I.~Guyon, U.~V. Luxburg, S.~Bengio, H.~Wallach, R.~Fergus, S.~Vishwanathan, and R.~Garnett, editors, \emph{Advances in NIPS}. Curran Associates, Inc., 2017.
\newblock URL \url{https://proceedings.neurips.cc/paper_files/paper/2017/file/3f5ee243547dee91fbd053c1c4a845aa-Paper.pdf}.

\bibitem[Waleffe et~al.(2024)Waleffe, Byeon, Riach, Norick, Korthikanti, Dao, Gu, Hatamizadeh, Singh, Narayanan, Kulshreshtha, Singh, Casper, Kautz, Shoeybi, and Catanzaro]{mamba2hybrid}
R.~Waleffe, W.~Byeon, D.~Riach, B.~Norick, V.~Korthikanti, T.~Dao, A.~Gu, A.~Hatamizadeh, S.~Singh, D.~Narayanan, G.~Kulshreshtha, V.~Singh, J.~Casper, J.~Kautz, M.~Shoeybi, and B.~Catanzaro.
\newblock An empirical study of mamba-based language models, 2024.
\newblock URL \url{https://arxiv.org/abs/2406.07887}.

\bibitem[Wang et~al.(2024)Wang, Paliotta, May, Rush, and Dao]{wang-2024-mamballama}
J.~Wang, D.~Paliotta, A.~May, A.~M. Rush, and T.~Dao.
\newblock The mamba in the llama: Distilling and accelerating hybrid models, 2024.
\newblock URL \url{https://arxiv.org/abs/2408.15237}.

\bibitem[Wen et~al.(2024)Wen, Dang, and Lyu]{Wen2024RNNsAN}
K.~Wen, X.~Dang, and K.~Lyu.
\newblock Rnns are not transformers (yet): The key bottleneck on in-context retrieval.
\newblock \emph{ArXiv}, abs/2402.18510, 2024.
\newblock URL \url{https://api.semanticscholar.org/CorpusID:268041425}.

\bibitem[Widrow and Hoff(1988)]{widrow-1988-adaptive}
B.~Widrow and M.~E. Hoff.
\newblock Adaptive switching circuits.
\newblock 1988.
\newblock URL \url{https://api.semanticscholar.org/CorpusID:60830585}.

\bibitem[Wu et~al.(2021)Wu, Wu, Daneshjou, Ouyang, Ho, and Zou]{wu-2021-fda}
E.~Wu, K.~Wu, R.~Daneshjou, D.~Ouyang, D.~E. Ho, and J.~Zou.
\newblock How medical {AI} devices are evaluated: limitations and recommendations from an analysis of {FDA} approvals.
\newblock \emph{Nature Medicine}, pages 582--584, 2021.
\newblock URL \url{https://doi.org/10.1038/s41591-021-01312-x}.

\bibitem[Xiong et~al.(2023)Xiong, Liu, Molybog, Zhang, Bhargava, Hou, Martin, Rungta, Sankararaman, Oguz, Khabsa, Fang, Mehdad, Narang, Malik, Fan, Bhosale, Edunov, Lewis, Wang, and Ma]{xiong-2023-llamalong}
W.~Xiong, J.~Liu, I.~Molybog, H.~Zhang, P.~Bhargava, R.~Hou, L.~Martin, R.~Rungta, K.~A. Sankararaman, B.~Oguz, M.~Khabsa, H.~Fang, Y.~Mehdad, S.~Narang, K.~Malik, A.~Fan, S.~Bhosale, S.~Edunov, M.~Lewis, S.~Wang, and H.~Ma.
\newblock Effective long-context scaling of foundation models, 2023.
\newblock URL \url{https://arxiv.org/abs/2309.16039}.

\bibitem[Yang and Zhang(2024)]{yang-2024-fla}
S.~Yang and Y.~Zhang.
\newblock {FLA: A Triton-Based Library for Hardware-Efficient Implementations of Linear Attention Mechanism}, 2024.
\newblock URL \url{https://github.com/sustcsonglin/flash-linear-attention}.

\bibitem[Yang et~al.(2024{\natexlab{a}})Yang, Wang, Shen, Panda, and Kim]{yang-etal-2024-gla}
S.~Yang, B.~Wang, Y.~Shen, R.~Panda, and Y.~Kim.
\newblock Gated linear attention transformers with hardware-efficient training.
\newblock In \emph{Proceedings of ICML}. PMLR, 2024{\natexlab{a}}.

\bibitem[Yang et~al.(2024{\natexlab{b}})Yang, Wang, Zhang, Shen, and Kim]{DBLP:journals/corr/abs-2406-06484}
S.~Yang, B.~Wang, Y.~Zhang, Y.~Shen, and Y.~Kim.
\newblock Parallelizing linear transformers with the delta rule over sequence length.
\newblock \emph{CoRR}, abs/2406.06484, 2024{\natexlab{b}}.
\newblock \doi{10.48550/ARXIV.2406.06484}.
\newblock URL \url{https://doi.org/10.48550/arXiv.2406.06484}.

\bibitem[Yang et~al.(2024{\natexlab{c}})Yang, Wang, Zhang, Shen, and Kim]{yang-2024-parallelizing}
S.~Yang, B.~Wang, Y.~Zhang, Y.~Shen, and Y.~Kim.
\newblock Parallelizing linear transformers with the delta rule over sequence length.
\newblock \emph{ArXiv}, abs/2406.06484, 2024{\natexlab{c}}.
\newblock URL \url{https://api.semanticscholar.org/CorpusID:270371554}.

\bibitem[Zellers et~al.(2019)Zellers, Holtzman, Bisk, Farhadi, and Choi]{zellers-2019-hellaswag}
R.~Zellers, A.~Holtzman, Y.~Bisk, A.~Farhadi, and Y.~Choi.
\newblock Hellaswag: Can a machine really finish your sentence?
\newblock In \emph{Proceedings of the 57th Annual Meeting of the Association for Computational Linguistics}, 2019.

\bibitem[Zhang et~al.(2022)Zhang, Jiang, Feng, Zheng, and Kong]{zhang-2022-cab}
J.~Zhang, S.~Jiang, J.~Feng, L.~Zheng, and L.~Kong.
\newblock Cab: Comprehensive attention benchmarking on long sequence modeling.
\newblock \emph{ArXiv}, abs/2210.07661, 2022.
\newblock URL \url{https://api.semanticscholar.org/CorpusID:252907545}.

\bibitem[Zhang et~al.(2024{\natexlab{a}})Zhang, Bhatia, Kumbong, and Ré]{zhang-2024-hedgehog}
M.~Zhang, K.~Bhatia, H.~Kumbong, and C.~Ré.
\newblock The hedgehog \& the porcupine: Expressive linear attentions with softmax mimicry, 2024{\natexlab{a}}.

\bibitem[Zhang et~al.(2024{\natexlab{b}})Zhang, Zeng, Wang, and Lu]{zhang-2024-tinyllama}
P.~Zhang, G.~Zeng, T.~Wang, and W.~Lu.
\newblock Tinyllama: An open-source small language model, 2024{\natexlab{b}}.

\bibitem[Zhang and Zhou(2017)]{zhang-2017-learning}
W.~Zhang and B.~Zhou.
\newblock Learning to update auto-associative memory in recurrent neural networks for improving sequence memorization.
\newblock \emph{ArXiv}, abs/1709.06493, 2017.
\newblock URL \url{https://api.semanticscholar.org/CorpusID:22458497}.

\bibitem[Zhang and Cai(2022)]{zhang-cai-2022-linearizing}
Y.~Zhang and D.~Cai.
\newblock Linearizing transformer with key-value memory.
\newblock In Y.~Goldberg, Z.~Kozareva, and Y.~Zhang, editors, \emph{Proceedings of EMNLP}, pages 346--359, Abu Dhabi, United Arab Emirates, 2022.
\newblock \doi{10.18653/v1/2022.emnlp-main.24}.
\newblock URL \url{https://aclanthology.org/2022.emnlp-main.24}.

\bibitem[Zhong et~al.(2021)Zhong, Yin, Yu, Zaidi, Mutuma, Jha, Awadallah, Celikyilmaz, Liu, Qiu, and Radev]{zhong-etal-2021-qmsum}
M.~Zhong, D.~Yin, T.~Yu, A.~Zaidi, M.~Mutuma, R.~Jha, A.~H. Awadallah, A.~Celikyilmaz, Y.~Liu, X.~Qiu, and D.~Radev.
\newblock {QMS}um: A new benchmark for query-based multi-domain meeting summarization.
\newblock In \emph{Proceedings of NAACL}, pages 5905--5921, Online, 2021.
\newblock \doi{10.18653/v1/2021.naacl-main.472}.
\newblock URL \url{https://aclanthology.org/2021.naacl-main.472}.

\bibitem[Zhou et~al.(2016)Zhou, Wu, Zhang, and Zhou]{zhou-2016-mgu}
G.-B. Zhou, J.~Wu, C.-L. Zhang, and Z.-H. Zhou.
\newblock Minimal gated unit for recurrent neural networks, 2016.

\end{thebibliography}
\bibliographystyle{abbrvnat}
\appendix

\newpage
\section{Linear Attention and its Chunkwise Form}
\label{appendix:la}
Linear Attention (LA)~\citep{katharopoulos-2020-transformers,qin-etal-2022-devil,qin-2024-transnormerllm} emerges as an alternative to resolve the quadratic complexity of self-attention (SA).
The key idea is to use the \emph{kernel trick}, which replaces $\operatorname{softmax}$ with a decomposable kernel function, resulting the following \emph{parallel form}:\footnote{
    There is a normalization term in vanilla LA similar to $\mathrm{softmax}$, \cite{qin-etal-2022-devil} reveal that removing it could avoid potential gradient explosions.
}
\begin{equation}
    \label{eq:linear-attention-parallel}
    \mathbf{O} = ((\phi(\mathbf{Q})\phi(\mathbf{K})^\top)\odot \mathbf{M})\mathbf{V}.
\end{equation}

where $\phi: \mathbb{R}^d \rightarrow \mathbb{R}^m$ functions as feature mapping applied to each input.
Unfolding Eq.~\ref{eq:linear-attention-parallel}, we have
\begin{equation}
    \begin{aligned}
        \boldsymbol{q}_t,\boldsymbol{k}_t,\boldsymbol{v}_t                                                                            & = \mathbf{W}_q \boldsymbol{x}_t, \mathbf{W}_k\boldsymbol{x}_t,\mathbf{W}_v\boldsymbol{x}_t\in\mathbb{R}^{d},                                                                                                                \\
        \boldsymbol{o}_t                                    = \sum_{i=1}^t \boldsymbol{v}_i  f(\boldsymbol{k}_i^\top\boldsymbol{q}_t) & =                \sum_{i=1}^t \boldsymbol{v}_i\phi(\boldsymbol{k}_i)^\top\phi(\boldsymbol{q}_t)   = \left[\mathbf{S}_t\equiv\sum_{i=1}^t \phi(\boldsymbol{k}_i)\otimes\boldsymbol{v}_i \right]^\top \phi(\boldsymbol{q}_t). \\
    \end{aligned}
\end{equation}
$\otimes$ means outer product operation.
It is clear that by leveraging the associativity, LA admits simple recurrent updating rules with matrix-valued hidden states $\mathbf{S}_t\in \mathbb{R}^{m\times d}$:
\begin{equation}
    \boldsymbol{o}_t  =  \mathbf{S}_t^\top \phi(\boldsymbol{q}_t);\;\mathbf{S}_t     = \mathbf{S}_{t-1} + \phi(\boldsymbol{k}_t) \otimes\boldsymbol{v}_t.
    \label{eq:linear-attention-recurrent}
\end{equation}
By reserving bounded $m$ memory slots only, the overall computation complexity is reduced from $O(T^2 d)$ to $O(T md)$.
When the sequence length is $T \gg m,d$, the $md$ factor has a minor impact on the complexity, and LA can be much more efficient than its counterpart with quadratic complexity.

During inference, LA enjoys the merits of RNNs, which only need to maintain $O(md)$ hidden memories, helping avoid the memory-cost KV cache management in SA mechanisms.
However, Eq.~\ref{eq:linear-attention-recurrent} employs a simple additive updating rule and can be hard to ``\emph{forget}'' unrelated information if necessary~\citep{peng-2021-rfa}, making the limited memory states vulnerable to be chaotic.

\paragraph{Gating mechanism} has played a key role in classical RNNs~\citep{hochreiter-1997-lstm,gers-1999-forget,cho-etal-2014-gru}, which serves as a mechanism to control the information flows in the network and help read and write from the memory selectively.
\cite{sun-2023-retnet} propose to apply a \emph{data-independent} gate to LA, significantly narrowing the gap between LA and SA: $\mathbf{S}_t = \lambda \mathbf{S}_{t-1}+\phi(\boldsymbol{k}_t)\otimes\boldsymbol{v}_t$, $\lambda \in[0,1]$ is a non-learnable scalar.
Recent work~\citep{yang-etal-2024-gla,katsch-2024-gateloop} further imposes a finer-grained \emph{data-dependent} gate:
\begin{equation}
    \label{eq:gla}
    \mathbf{S}_t = \mathrm{Diag}(\boldsymbol{\alpha}_t) \mathbf{S}_{t-1} + \phi(\boldsymbol{k}_t)\otimes\boldsymbol{v}_t,
\end{equation}
where each $\boldsymbol{\alpha}_t\in [0,1]^{m}$ from $\mathbf{A} := \{\boldsymbol{\alpha}_i\}_{i=1}^T\in [0,1]^{T\times m}$ is dependent on the input.
Alternatively, we can couple the key values with the forget gates by allowing $\phi(\boldsymbol{k}_t) = 1-\boldsymbol{\alpha}_t$ in spirit of \cite{cho-etal-2014-gru,zhou-2016-mgu} and \cite{qin-2024-hgrn2}, which reduces the number of parameters and improves efficiency accordingly.

\subsection{Hardware-Efficient Training}
Despite the theoretical advantages of linear complexity, the recurrent form of Eq.~\ref{eq:linear-attention-recurrent} is still inefficient during training.
Such recurrent computation prevents the full utilization of modern GPU parallelism over sequence lengths \citep{martin-2018-parallelizing,rush-2020-torch}.
On the other hand, the parallel form (Eq.~\ref{eq:linear-attention-parallel}) can be parallelized in similar vein as in flash attention \citep{dao-2022-flashattn,dao-2024-flashattn2}.
However, due to the existence of the casual mask $\mathbf{M}$, we can not rearrange its computation order by $\mathbf{K}\mathbf{V}$ first, so that the parallel form still adheres to the quadratic complexity, which can hardly be scaled to very-long training context (e.g., sequences with more than 8K tokens).

\paragraph{Chunkwise form} recurrences have been carried forward by \cite{sun-2023-retnet}, and achieve a good trade-off between the recurrent and parallel forms.
\cite{yang-etal-2024-gla} further disclose that the element-wise gating of Eq.~\ref{eq:gla} also satisfies the associative property required by parallel scan
\citep{blelloch-1993-prefix} and derive a parallelized chunkwise gated linear attention in a similar vein.
The key idea is to partition the sequence into $N=\lceil \frac{T}{C} \rceil$ chunks of size $C$ with $\mathbf{Q}_{[t]}=\boldsymbol{q}_{tC}, \boldsymbol{q}_{tC+1},\ldots,\boldsymbol{q}_{tC+C}$, and so forth for $\mathbf{K}_{[t]},\mathbf{V}_{[t]}\in \mathbb{R}^{C\times d},\mathbf{A}_{[t]}\in \mathbb{R}^{C\times m}$.
Firstly, unrolling the $i$-th hidden state in the $t$-th chunk in Eq.~\ref{eq:gla}, we get
\begin{equation}
    \label{eq:gla-recurrent}
    \begin{aligned}
        \mathbf{S}_{[t],i} & = \mathrm{Diag}\left(\mathbf{A}_{[t],i}\right)\mathbf{S}_{[t],i-1}+ \phi\left(\mathbf{K}_{[t],i}\right)\otimes\mathbf{V}_{[t],i}  = \cdots \\
                           & =  \mathrm{Diag}\left(\prod_{j=1}^{i} \mathbf{A}_{[t],j}\right)\mathbf{S}_{[t-1],C}
        + \sum_{k=1}^{i}\left(\phi(\mathbf{K}_{[t],k})\odot\prod_{j=k+1}^{i} \mathbf{A}_{[t],j}\right)\otimes \mathbf{V}_{[t],k}
    \end{aligned}
\end{equation} %
We write the last hidden in the chunk $\mathbf{S}_{[t],C}$ as $\mathbf{S}_{[t]}$ interchangeably for simplicity.
Define $\overrightarrow{\mathcal{A}}_{[t],i}=\prod_{j=1}^{i} \mathbf{A}_{[t],j}\in [0,1]^d$ as the cumulative decay from the start of chunk to $i$, and likewise
$\overleftarrow{\mathcal{A}}_{[t],i}=\prod_{j=i+1}^{C} \mathbf{A}_{[t],j}\in [0,1]^d$ from $i+1$ to the end of the chunk, then
\begin{equation}
    \label{eq:gla-chunk-gate}
    \begin{aligned}
        \mathbf{S}_{[t]} & = \mathrm{Diag}(\overrightarrow{\mathcal{A}}_{[t],C})\mathbf{S}_{[t-1]}
        + (\mathbf{K}_{[t]}\odot\overleftarrow{\mathcal{A}}_{[t]})^\top \mathbf{V}_{[t]}
    \end{aligned}
\end{equation}
$\overrightarrow{\mathcal{A}},\overleftarrow{\mathcal{A}}$ can be absorbed into $\mathbf{Q,K}$ first : $\overline{\mathbf{Q}}_{[t]}=\phi(\mathbf{Q}_{[t]})\odot\overrightarrow{\mathcal{A}}_{[t]}$,
$\overline{\mathbf{K}}_{[t]}=\phi(\mathbf{K}_{[t]})\odot(\overleftarrow{\mathcal{A}}_{[t]}/\overrightarrow{\mathcal{A}}_{[t],C})$.
Combining them with Eq.~\ref{eq:linear-attention-parallel} and Eq.~\ref{eq:gla-recurrent}, we derive the following vectorized updating rules
\begin{equation}
    \label{eq:gla-chunk}
    \begin{aligned}
        \mathbf{O}_{[t]} =\overline{\mathbf{Q}}_{[t]}\mathbf{S}_{[t-1]}
        + \left(\overline{\mathbf{Q}}_{[t]} \overline{\mathbf{K}}_{[t]}^\top \odot\mathbf{M}_{[t]}\right) \mathbf{V}_{[t]}
    \end{aligned}
\end{equation}
The first term is referred to as the \emph{inter} chunk part and the second term is the \emph{intra} chunk part.
The process to get this \emph{intra} part is a little more involved as the cumulative productions of $\overleftarrow{\mathcal{A}}_{[t]}/\overrightarrow{\mathcal{A}}_{[t],C}$ is greater than 1, which can lead to numerical instability.
\cite{yang-etal-2024-gla} deal with this issue by proposing a secondary-chunking strategy, and we refer readers to their paper for more details.

\paragraph{Hardward considerations}
Modern GPU architectures, such as the NVIDIA A100, offer highly optimized matrix multiplication (matmul) operations through specialized Tensor Cores, achieving up to 16$\times$ higher throughput than non-matmul operations~\citep{dao-2022-flashattn}.
However, this incurs IO overheads due to data transfer from slower, off-chip global high bandwidth memory (HBM) to on-chip shared memory (SRAM).
The chunkwise form balances I/O and computation complexity tradeoffs.
As shown in Eq.\ref{eq:gla-chunk}, it improves parallelism over the sequence dimension while reducing non-matmul FLOPs greatly.
Also, the chunk recurrent updating conducts the query and hidden states reduction in an online manner, requiring only $O(N dm)$ hidden states materialized into HBMs, so that it can significantly reduce the memory/IO overheads.
While LA enjoys much lower overall running FLOPs than SA, the chunkwise form displays a practical significant wall-clock speedup against SA, due to its hardware-efficient implementations~\citep{
    yang-2024-fla}.

\newcommand*{\obox}{\tcboxmath[colback=limegreen!40, colframe=limegreen!40, valign=center,height=18pt,size=fbox,arc=0pt,box align=base,baseline=2.5mm]}
\newcommand*{\qbox}{\tcboxmath[colback=blue!15, colframe=blue!15, valign=center,height=18pt,size=fbox, arc=0pt,box align=base,baseline=2.5mm]}
\newcommand*{\shbox}{\tcboxmath[colback=orange!20, colframe=orange!20, valign=center,height=18pt,size=fbox, arc=0pt,box align=base,baseline=2.5mm]}
\newcommand*{\kvbox}{\tcboxmath[colback=red!20, colframe=red!20, valign=center,height=18pt,size=fbox, arc=0pt,box align=base,baseline=2.5mm]}

\begin{algorithm*}[t!]
  \tiny
  \begin{multicols}{2}

    \begin{algorithmic}[1]
      \Def \textsc{ForwardPass}($\mathbf{Q,K,V,I,A}$)
      \State Divide $\mathbf{Q,K,V}\in \R^{T \times d},\mathbf{I,A}\in\R^{T \times m}$
      \State \qquad into $N = \left\lceil \frac{T}{C}\right\rceil$ blocks\Comment{$C$ is chunk size}

      \Function{$\mathtt{chunk\_cumsum}$}{$\mathbf{A}$}
      \ParFor{$n \gets 1, N$}
      \State Load $\mathbf{A}_{[n]}$ to SRAM
      \State Store $\overrightarrow{\mathcal{A}}_{[n]} \leftarrow \mathtt{cumsum}(\mathbf{A}_{[n]})$ to HBM
      \EndParFor
      \State \Return $\overrightarrow{\mathcal{A}}\gets \overrightarrow{\mathcal{A}}_{[0]},\dots,\overrightarrow{\mathcal{A}}_{[N]}$
      \EndFunction

      \Statex
      \Function{$\mathtt{gsa\_fwd}$}{$\mathbf{Q},\mathbf{K},\mathbf{V},\overrightarrow{\mathcal{A}}, \mathtt{GATE\_K}$}

      \State On chip: construct causal mask $\mathbf{M}\in \R^{C\times C}$
      \For{$n \gets 1, N$}
      \State{Store $\mathbf{S}$ to HBM as $\mathbf{S}_{[n]}$}\Comment{Initialize $\mathbf{S} = \bm{0}$}
      \State{Load $\mathbf{K}_{[n]}$, $\mathbf{V}_{[n]},\overrightarrow{\mathcal{A}}_{[n],C}\overleftarrow{\mathcal{A}}_{[n]}$ to SRAM}
      \State On chip: $\overleftarrow{\mathcal{A}}_{[n]} \leftarrow \overrightarrow{\mathcal{A}}_{[n],C}/\overrightarrow{\mathcal{A}}_{[n]}$
      \If{$\mathtt{GATE\_K}$}
      \State $\mathbf{S} \gets \mathrm{Diag}(\overrightarrow{\mathcal{A}}_{[n],C})\mathbf{S}
        + (\mathbf{K}_{[n]}\odot\overleftarrow{\mathcal{A}}_{[n]})^\top \mathbf{V}_{[n]}$
      \Else
      \State $\mathbf{S} \gets \mathbf{S}\mathrm{Diag}(\overrightarrow{\mathcal{A}}_{[n],C})
        + \mathbf{K}_{[n]}^\top (\mathbf{V}_{[n]}\odot\overleftarrow{\mathcal{A}}_{[n]})$
      \EndIf
      \color{black}
      \EndFor
      \ParFor{$n \gets 1, N$}
      \State Load $\mathbf{Q}_{[n]}, \mathbf{K}_{[n]}, \mathbf{V}_{[n]}, \mathbf{S}_{[n]},\overleftarrow{\mathcal{A}}_{[n]},\overrightarrow{\mathcal{A}}_{[n]}$ to SRAM
      \State On chip:
      \quad\If{$\mathtt{GATE\_K}$}
      \State $\bar{\mathbf{Q}}_{[n]}\gets\mathbf{Q}_{[n]}\odot\overrightarrow{\mathcal{A}}_{[n]}$
      \State $\bar{\mathbf{K}}_{[n]}\gets\mathbf{K}_{[n]}\odot(\overleftarrow{\mathcal{A}}_{[n]}/\overrightarrow{\mathcal{A}}_{[n],C})$
      \State $\mathbf{O}_{[n]} \gets\bar{\mathbf{Q}}_{[n]}\mathbf{S}_{[n-1]}
        + \left(\mathbf{P}\equiv\bar{\mathbf{Q}}_{[n]} \bar{\mathbf{K}}_{[n]}^\top \odot\mathbf{M}\right) \mathbf{V}_{[n]}$
      \Else
      \State $\bar{\mathbf{V}}_{[n]}\gets\mathbf{V}_{[n]}\odot(\overleftarrow{\mathcal{A}}_{[n]}/\overrightarrow{\mathcal{A}}_{[n],C})$
      \State $\mathbf{O}_{[n]} \gets\mathbf{Q}_{[n]}\mathbf{S}_{[n-1]} + \left(\mathbf{P}\equiv\mathbf{Q}_{[n]} {\mathbf{K}}_{[n]}^\top \odot\mathbf{M}\right) \bar{\mathbf{V}}_{[n]}$
      \State $\mathbf{O}_{[n]} \gets\mathbf{O}_{[n]}\odot\overrightarrow{\mathcal{A}}_{[n]}$
      \EndIf
      \State Store $\mathbf{O}$ to HBM as $\mathbf{O}_{[n]}$.
      \EndParFor
      \State \Return $\mathbf{O}_{[1,\dots,N]},\mathbf{S}_{[1,\dots,N]}$
      \EndFunction

      \Statex
      \State $\overrightarrow{\mathcal{A}} \gets \mathtt{chunk\_cumsum}(\mathbf{A})$  \Comment{preprocessing}
      \State $\mathbf{O}^k,\mathbf{S}^k \gets \mathtt{gsa\_fwd}(\mathbf{Q},\mathbf{K},\mathbf{I},\overrightarrow{\mathcal{A}},\mathtt{False})$
      \State $\mathbf{Q}^v\gets \operatorname{softmax}(\mathbf{O}^k)$
      \State $\mathbf{O},\mathbf{S}^v \gets \mathtt{gsa\_fwd}(\mathbf{Q}^v,\mathbf{I},\mathbf{V},\overrightarrow{\mathcal{A}},\mathtt{True})$
      \State \Return $\mathbf{O}$

    \end{algorithmic}

    \columnbreak

    \begin{algorithmic}[2]
      \Def \textsc{BackwardPass}($\mathbf{Q,K,V,I,A},\mathbf{O}^k,\mathrm{d}\mathbf{O}$)
      \State Divide $\mathbf{Q,K,V,O},\mathrm{d}\mathbf{O}\in \R^{T \times d},\mathbf{I,A}\in\R^{T \times m}$
      \State \qquad into $N = \left\lceil \frac{T}{C}\right\rceil$ blocks\Comment{$C$ is chunk size}
      \Function{$\mathtt{gsa\_bwd}$}{$\mathbf{Q},\mathbf{K},\mathbf{V},\mathbf{S},\overrightarrow{\mathcal{A}},\mathrm{d}\mathbf{O}, \mathtt{GATE\_K}$}

      \State On chip: construct causal mask $\mathbf{M}\in \R^{C\times C}$ %

      \For{$n \gets N, 1$}   \Comment{in reverse order}
      \State Store $\mathrm{d}\mathbf{S}$ in HBM as $\mathrm{d}\mathbf{S}_{[n]}$\Comment{Initialize $\mathrm{d}\mathbf{S} = \bm{0}$}
      \State{Load $\mathbf{Q}_{[n]}, \overrightarrow{\mathcal{A}}_{[n]}, \mathrm{d}\mathbf{O}_{[n]}$ to SRAM}
      \State On chip:
      \If{$\mathtt{GATE\_K}$}
      \State \hspace{-8pt}$\mathrm{d}\mathbf{S} \gets \mathrm{Diag}(\overrightarrow{\mathcal{A}}_{[i],C})\mathrm{d}\mathbf{S} + (\mathbf{Q}_{[n]}\odot\overrightarrow{\mathcal{A}}_{[n]})^\top \mathrm{d}\mathbf{O}_{[n]}$
      \Else
      \State\hspace{-8pt}$\mathrm{d}\mathbf{S} \gets \mathrm{d}\mathbf{S}\mathrm{Diag}(\overrightarrow{\mathcal{A}}_{[i],C}) + \mathbf{Q}_{[n]}^\top (\mathrm{d}\mathbf{O}_{[n]}\odot\overrightarrow{\mathcal{A}}_{[n]})$
      \EndIf

      \EndFor
      \ParFor{$n \gets 1, N$}
      \State{Load $\mathbf{Q}_{[n]}, \mathbf{K}_{[n]},\mathbf{V}_{[n]}, \mathrm{d}\mathbf{O}_{[n]} \in \R^{C \times d}$ }
      \State{\quad$\mathbf{S}_{[n]}$, $\mathrm{d}\mathbf{S}_{[n]} \in \R^{d \times d}$ to SRAM}
      \State On chip:\Comment{Recompute $\overleftarrow{\mathcal{A}}_{[n]},\bar{\mathbf{Q}}_{[n]},\bar{\mathbf{K}}_{[n]},\bar{\mathbf{V}}_{[n]},\mathbf{P}$}
      \If{$\mathtt{GATE\_K}$}%
      \State $\mathrm{d}\mathbf{P} \leftarrow (\mathrm{d}\mathbf{O}_{[n]} \mathbf{V}_{[n]}^{\top}) \odot \mathbf{M}$
      \State $\mathrm{d}\mathbf{Q} \leftarrow  (\mathrm{d}\mathbf{O}_{[n]} \mathbf{S}+ \mathrm{d}\mathbf{P}\bar{\mathbf{K}}_{[n]}^\top)\odot\overrightarrow{\mathcal{A}}_{[n]}$
      \State $\mathrm{d}\mathbf{K} \leftarrow  ( \mathbf{V}_{[n]} \mathrm{d}\mathbf{S}^\top+\mathrm{d}\mathbf{P}^\top\bar{\mathbf{Q}}_{[n]})\odot\overleftarrow{\mathcal{A}}_{[n]}$
      \State $\mathrm{d}\mathbf{V} \leftarrow  \bar{\mathbf{K}}_{[n]} \mathrm{d}\mathbf{S}_{[n]} + \mathbf{P}^\top  \mathrm{d}\mathbf{O}_{[n]}$
      \Else
      \State $\mathrm{d}\mathbf{P} \leftarrow (\mathrm{d}\mathbf{O}_{[n]} \bar{\mathbf{V}}_{[n]}^{\top}) \odot \mathbf{M}$
      \State $\mathrm{d}\mathbf{Q} \leftarrow  \mathrm{d}\mathbf{O}_{[n]} \mathbf{S}^\top+ \mathrm{d}\mathbf{P}\mathbf{K}_{[n]}$
      \State $\mathrm{d}\mathbf{K} \leftarrow  \bar{\mathbf{V}}_{[n]} \mathrm{d}\mathbf{S}^\top+\mathrm{d}\mathbf{P}^\top\mathbf{Q}_{[n]}$
      \State $\mathrm{d}\mathbf{V} \leftarrow (\mathbf{K}_{[n]} \mathrm{d}\mathbf{S}_{[n]} + \mathbf{P}^\top  \mathrm{d}\mathbf{O}_{[n]})\odot\overleftarrow{\mathcal{A}}_{[n]}$
      \EndIf
      \State Write $ \mathrm{d}\mathbf{Q}, \mathrm{d}\mathbf{K}, \mathrm{d}\mathbf{V}$ to HBM as $\mathrm{d}\mathbf{Q}_{[n]}, \mathrm{d}\mathbf{K}_{[n]}, \mathrm{d}\mathbf{V}_{[n]}$
      \EndParFor
      \color{black}

      \State \Return $\mathrm{d}\mathbf{Q}_{[1,\dots,N]},\mathrm{d}\mathbf{K}_{[1,\dots,N]},\mathrm{d}\mathbf{V}_{[1,\dots,N]}$
      \EndFunction

      \Statex
      \State Recompute $\overrightarrow{\mathcal{A}},\mathbf{S}^k,\mathbf{S}^v$
      \State $\mathrm{d}\mathbf{Q}^v,\mathrm{d}\mathbf{I}^v,\mathrm{d}\mathbf{V}\gets \mathtt{gsa\_bwd}(\mathbf{Q},\mathbf{I},\mathbf{V},\mathbf{S}^v,\overrightarrow{\mathcal{A}},\mathrm{d}\mathbf{O},\mathtt{False})$
      \State $\mathrm{d}\mathbf{O}^k\gets \mathrm{d}\operatorname{softmax}(\mathbf{O}^k,\mathrm{d}\mathbf{Q}^v)$\Comment{$\operatorname{softmax}$ gradients}
      \State $\mathrm{d}\mathbf{Q},\mathrm{d}\mathbf{K},\mathrm{d}\mathbf{I}^k\gets \mathtt{gsa\_bwd}(\mathbf{Q},\mathbf{K},\mathbf{I},\mathbf{S}^k,\overrightarrow{\mathcal{A}},\mathrm{d}\mathbf{O}^k,\mathtt{True})$
      \State $\mathrm{d}\mathbf{I}\gets \mathrm{d}\mathbf{I}^k+\mathrm{d}\mathbf{I}^v$
      \State $\mathrm{d}\mathbf{A}\gets \mathtt{reversed\_cumsum}(\mathbf{Q}\odot\mathrm{d}\mathbf{Q}-\mathbf{K}\odot\mathrm{d}\mathbf{K}+$
        \State \qquad\qquad\qquad\qquad\qquad\qquad $\mathbf{O}\odot\mathrm{d}\mathbf{O}-\mathbf{V}\odot\mathrm{d}\mathbf{V})$
      \State \Return $\mathrm{d}\mathbf{Q},\mathrm{d}\mathbf{K},\mathrm{d}\mathbf{V},\mathrm{d}\mathbf{I},\mathrm{d}\mathbf{A}$
    \end{algorithmic}
  \end{multicols}
  \caption{\small Hardware-Efficient Gated Slot Attention}
  \label{alg:gsa}
\end{algorithm*}

\section{Details for \textsc{GSA}}\label{appendix:alg}

\begin{wrapfigure}[16]{r}{0.58\textwidth}
  \vspace{-3\normalbaselineskip}
  \centering
  \begin{tikzpicture}[
    qnode/.style={
        draw,
        minimum width=19pt,
        minimum height=19pt,
        very thick,
        rounded corners=2pt,
        fill=blue!15,
        font=\tiny,
      },
    kvnode/.style={
        draw,
        minimum width=24pt,
        minimum height=19pt,
        very thick,
        rounded corners=2pt,
        fill=red!20,
        font=\tiny,
      },
    qkvnode/.style={
        draw,
        dashed,
        thick,
        inner sep=2pt,
        minimum width=19pt,
        minimum height=19pt,
        rounded corners=2pt,
        font=\tiny,
      },
    onode/.style={
        draw,
        minimum width=19pt,
        minimum height=19pt,
        very thick,
        rounded corners=2pt,
        fill=limegreen!40,
        font=\tiny,
      },
    hidden/.style={
        draw,
        minimum width=19pt,
        minimum height=19pt,
        very thick,
        rounded corners=2pt,
        font=\tiny,
      },
    hiddenlink/.style={
    dashed,
    ->,>={Straight Barb[width=4pt]},
    thick,
    shorten >=3pt,
    shorten <=3pt,
    },
    holink/.style={
    ->,>={Straight Barb[width=4pt]},
    thick,
    shorten <=2pt,shorten >=2pt,
    },
    qkvlink/.style={
    ->,>={Straight Barb[width=4pt]},
    thick,
    shorten >=2pt,
    },
    olink/.style={
    ->,>={Straight Barb[width=4pt]},
    thick,
    },
    ]

    \foreach \i in {1,...,2}{
    \node[qnode,anchor=west] (i\i) at (\i*3.5+0.5,0) {$\bar{\mathbf{Q}}_{[\i]}$};
    \node[kvnode,anchor=west] (kv0\i) at ([xshift=0pt]i\i.east) {$\mathbf{K}_{[\i]};\mathbf{I}_{[\i]}$};
    \node[qkvnode, fit=(i\i)(kv0\i)] (qkv0\i) {};
    }
    \foreach \i in {1,...,3}{
    \tikzmath{\j=int(\i-1); }
    \node[hidden, fill=orange!\the\numexpr\i*20-20] (h0\i) at (\i*3.5,30pt) {$\mathbf{S}^k_{[\j]}$};
    \ifnum\i>1
      \draw[hiddenlink] (h0\the\numexpr\i-1) -- (h0\i);
      \draw[olink] (kv0\the\numexpr\i-1) -- (h0\i);
    \fi
    }
    \foreach \i in {1,...,2}{
    \node[onode,anchor=west] (o0\i) at (\i*3.5+0.5,60pt) {$\mathbf{O}^{k}_{[\i]}$};
    \node[kvnode,anchor=west] (kv1\i) at ([xshift=0pt]o0\i.east) {$\mathbf{I}_{[\i]};\bar{\mathbf{V}}_{[\i]}$};
    \node[qkvnode, fit=(o0\i)(kv1\i)] (qkv1\i) {};
    \draw[qkvlink] (qkv0\i) -- (o0\i);
    \draw[holink] (h0\the\numexpr\i) -- (o0\i);
    }
    \foreach \i in {1,...,3}{
    \tikzmath{\j=int(\i-1); }
    \node[hidden, fill=orange!\the\numexpr\i*20-20] (h1\i) at (\i*3.5,90pt) {$\mathbf{S}^v_{[\j]}$};
    \ifnum\i>1
      \draw[hiddenlink] (h1\the\numexpr\i-1) -- (h1\i);
      \draw[olink] (kv1\the\numexpr\i-1) -- (h1\i);
    \fi

    }
    \foreach \i in {1,...,2}{
    \node[onode,anchor=west] (o1\i) at (\i*3.5+0.5,120pt) {$\mathbf{O}_{[\i]}$};
    \draw[qkvlink] (qkv1\i) -- (o1\i);
    \draw[holink] (h1\the\numexpr\i) -- (o1\i);
    key      }

  \end{tikzpicture}
  \caption{\small
    Diagrams of the recurrence and updating rules in Gated Slot Attention.
    The outputs of the first pass is taken as queries of the second pass.\\
    \tcbox[colback=blue!20,arc=1pt,boxrule=1pt,top=0pt,bottom=0pt,left=1pt,right=1pt,width=10pt,height=10pt,nobeforeafter,tcbox raise=-2pt]{}: query nodes \qquad
    \tcbox[colback=red!20,arc=1pt,boxrule=1pt,top=0pt,bottom=0pt,left=1pt,right=10pt,width=10pt,height=10pt,nobeforeafter,tcbox raise=-2pt]{}: key/value nodes\qquad\\
    \tcbox[colback=limegreen!40,arc=1pt,boxrule=1pt,top=0pt,bottom=0pt,left=1pt,right=1pt,width=10pt,height=10pt,nobeforeafter,tcbox raise=-2pt]{}: output nodes\qquad
    \tcbox[colback=orange!40,arc=1pt,boxrule=1pt,top=0pt,bottom=0pt,left=1pt,right=1pt,width=10pt,height=10pt,nobeforeafter,tcbox raise=-2pt]{}: recurrent hidden states
  }
  \label{fig:gsa-2pass}
\end{wrapfigure}

Beyond the recurrent GSA form provided in Figure.~\ref{fig:gsa-recurrent}, we give detailed, hardware-efficient procedures for the forward and backward passes of Gated Slot Attention (\textsc{GSA}) in Algorithm~\ref{alg:gsa}.
For simplicity, we define $\mathbf{A} = \{\boldsymbol{\alpha}_i\}_{i=1}^T\in [0,1]^{T\times m}$, and $\mathbf{I} = \{\boldsymbol{1}-\boldsymbol{\alpha}_i\}_{i=1}^T\in [0,1]^{T\times m}$.
The algorithm demonstrates that \textsc{GSA} can be modeled as a two-pass GLA, as illustrated in Fig.~\ref{fig:gsa-2pass}.

In the preprocessing step, we pre-compute the chunkwise cumulative sum of the forget gate, resulting in $\overrightarrow{\mathcal{A}}$.
Subsequently, $\overrightarrow{\mathcal{A}}$ along with the queries, keys, and values are passed to engage in two GLA passes.
For each chunk of size $C$, we define $\overleftarrow{\mathcal{A}}_{[i]} := \overrightarrow{\mathcal{A}}_{[i],C}/\overrightarrow{\mathcal{A}}_{[i]}$ as in Eq.~\ref{eq:gla-chunk-gate} and Eq.~\ref{eq:gla-chunk}.

In the first pass, $\overrightarrow{\mathcal{A}},\overleftarrow{\mathcal{A}}$ is absorbed into $\mathbf{Q,K}$ : $\bar{\mathbf{Q}}_{[i]}=\mathbf{Q}_{[i]}\odot\overrightarrow{\mathcal{A}}_{[i]}$,
$\bar{\mathbf{K}}_{[i]}=\mathbf{K}_{[i]}\odot(\overleftarrow{\mathcal{A}}_{[i]}/\overrightarrow{\mathcal{A}}_{[i],C})$,
then $\bar{\mathbf{Q}}$ and $\bar{\mathbf{K}}$ function as usual queries and keys, and the slot representations $\mathbf{I}$ serve as the value vectors.
\begin{equation*}
  \obox{\mathbf{O}_{[i]}^k} = \underbrace{\qbox{\bar{\mathbf{Q}}^{\textcolor{blue!15}{k}}_{[i]}}\shbox{\mathbf{S}^k_{[i-1]}}}_{\mathbf{O}^{\mathrm{inter}}_{[i]}} + \underbrace{((\qbox{\bar{\mathbf{Q}}^{\textcolor{blue!15}{k}}_{[i]}}\kvbox{\bar{\mathbf{K}}^{\top}_{[i]})\odot \mathbf{M})\mathbf{I}_{[i]}}}_{\mathbf{O}^{\mathrm{intra}}_{[i]}} \in \mathbb{R}^{C\times m}
\end{equation*}
We use different notations from those presented in Eq.\ref{eq:gsa} to enhance clarity in the chunkwise updating rules.
The output $\mathbf{O}^k$ is decomposed into the \emph{inter-chunk} recurrence and \emph{intra-chunk} parallel computations \citep{yang-etal-2024-gla}.

In the second pass, the output $\mathbf{O}^k$ from the first pass, after the application of the $\operatorname{softmax}$ function, serves as the queries $\mathbf{Q}^v$,
\begin{equation*}
  \qbox{\mathbf{Q}^{v}_{\textcolor{blue!15}{i}}} = \operatorname{softmax}(\qbox{\mathbf{O}^k_{\textcolor{blue!15}{i}}})
\end{equation*}
and $\mathbf{I}$/$\mathbf{V}$ are used as the key/value vectors, respectively.
The final \textsc{GSA} output $\mathbf{O}$ is obtained as follows:
\begin{equation*}
  \obox{\mathbf{O}^{\textcolor{limegreen!40}{v}}_{[i]}} = \qbox{\mathbf{Q}^{v}_{[i]}}\shbox{\mathbf{S}^v_{[i-1]}} + ((\qbox{\mathbf{Q}^{v}_{[i]}}\kvbox{\mathbf{I}^{\top}_{[i]})\odot \mathbf{M})\bar{\mathbf{V}}_{[i]}} \in \mathbb{R}^{C\times d}
\end{equation*}
Unlike in the first pass, $\overrightarrow{\mathcal{A}},\overleftarrow{\mathcal{A}}$ is absorbed into $\mathbf{V,O}$ rather than $\mathbf{Q,K}$.

During the backward pass, computing the gradients of $\mathbf{Q,K,V,I,A}$ involves variables already computed in the forward pass.
However, directly saving all intermediate results can pose severe challenges for memory management.
To address this issue, we adopt gradient checkpointing \cite{chen-2016-training} to trade off memory consumption for recomputation.
In addition to the input $\mathbf{Q,K,V,I,A}$, we selectively save only the output of the first GLA pass, which significantly reduces memory consumption (Figure~\ref{fig:memory}).

Similar to the forward pass, the backward pass involves two GLA backward passes as well, but in the reverse order.
The final gradient $\mathrm{d}\mathbf{I}$ is obtained by combining the gradients from these computations, i.e., $\mathrm{d}\mathbf{I} = \mathrm{d}\mathbf{I}^k + \mathrm{d}\mathbf{I}^v$.
The forget gate gradient can be decomposed into two parts: $\mathbf{Q}\odot\mathrm{d}\mathbf{Q} - \mathbf{K}\odot\mathrm{d}\mathbf{K}$ and $\mathbf{O}\odot\mathrm{d}\mathbf{O} - \mathbf{V}\odot\mathrm{d}\mathbf{V}$ (cf. \S C in \cite{yang-etal-2024-gla}).
The reversed cumulative sum in the backward pass corresponds to the cumulative sum computed in the preprocessing step of the forward pass.

We provide a PyTorch implementation for the above algorithm with chunkwise parallelism in Listing~\ref{listing:code}.

\newenvironment{longlisting}{\captionsetup{type=listing,labelfont=bf}}{}
\renewcommand{\theFancyVerbLine}{\ttfamily\textcolor[rgb]{0.5,0.5,0.5}{\scriptsize\arabic{FancyVerbLine}}}
\begin{longlisting}
    \begin{minted}[
    mathescape,
    linenos,
    numbersep=4pt,
    frame=lines,
    fontsize=\small,
    framesep=4mm,
]{python}
def gsa_fwd_k(q, k, v, g, C):
    '''
    q/k/v:
         query, key, value of shape [NC, C, K|V]
    g:
        local cumulative product of forget gate in log space
    C:
        chunk size
    '''
    # NC: number of chunks
    # K: query/key head dimension
    # V: value head dimension
    NC, C, K, V = *q.shape, v.shape[-1]
    # [K, V]
    s = q.new_zeros(K, V)
    # [NC, C, V]
    o = torch.empty_like(v)

    for i in range(0, NC):
        # [C, K|V] chunking
        c_q, c_k, c_v, c_g = q[i], k[i], v[i], g[i]
        # the last g of each chunk
        c_gn = c_g[-1]
        # inter-chunk w/ matmul
        c_vg, c_gn = c_v * (c_gn - c_g).exp(), c_gn.exp()
        # [C, V]
        c_o_inter = (c_q @ s) * c_g.exp()
        # hidden state update
        s = c_gn * s + c_k.t() @ c_vg

        # intra-chunk
        # [C, C]
        c_A = c_q @ c_k.t()
        # [C, V]
        c_o_intra = torch.zeros_like(c_v)
        for j in range(0, C // 16):
            t = slice(j * 16, j * 16 + 16)
            # [16, K|V] subchunking
            s_A, s_v, s_g = c_A[t], c_v[t], c_g[t]
            s_o = q.new_zeros(16, V)

            # inter-subchunk w/ matmul
            s_gn = s_g[0]
            for si in range(0, j):
                u = slice(si * 16, si * 16 + 16)
                s_o += s_A[:, u] @ (c_v[u] * (s_gn - c_g[u]).exp())
            s_o *= (s_g - s_gn).exp()
            # intra-subchunk w/o matmul
            for si in range(16):
                for sj in range(si + 1):
                    s_o[si] += s_A[si, j * 16 + sj] * s_v[sj] * (s_g[si] - s_g[sj]).exp()
            c_o_intra[t] = s_o
        # [C, V]
        o[i] = c_o_inter + c_o_intra
    return o


def gsa_fwd_v(q, k, v, g, C):
    NC, C, K, V = *q.shape, v.shape[-1]
    s = q.new_zeros(K, V)
    o = torch.empty_like(v)

    for i in range(0, NC):
        # [C, K|V] chunking
        c_q, c_k, c_v, c_g = q[i], k[i], v[i], g[i]
        # the last g of each chunk
        c_gn = c_g[-1]
        # inter-chunk w/ matmul
        c_qg, c_kg, c_gn = c_q * c_g.exp(), c_k * (c_gn - c_g).exp(), c_gn.exp()
        # [C, V]
        c_o_inter = c_qg @ s
        # hidden state update
        s = c_gn[:, None] * s + c_kg.t() @ c_v

        # intra-chunk
        c_A = q.new_zeros(C, C)
        for j in range(0, C // 16):
            t = slice(j * 16, j * 16 + 16)
            # [16, K|V] subchunking
            s_q, s_k, s_g = c_q[t], c_k[t], c_g[t]
            s_A = q.new_zeros(16, 16)

            # intra-subchunk w/o matmul
            for si in range(16):
                for sj in range(si + 1):
                    s_A[si, sj] = torch.sum(s_q[si] * s_k[sj] * (s_g[si] - s_g[sj]).exp())
            c_A[t, t] = s_A
            # inter-subchunk w/ matmul
            s_gn = s_g[0]
            s_qg = s_q * (s_g - s_gn).exp()
            for si in range(0, j):
                u = slice(si * 16, si * 16 + 16)
                c_A[t, u] = s_qg @ (c_k[u] * (s_gn - c_g[u]).exp()).t()
        c_o_intra = c_A @ c_v
        # [C, V]
        o[i] = c_o_inter + c_o_intra
    return o


def gsa(q, k, v, s, g):
    T, M = s.shape
    # reshape each input to [NC, C, K|V]
    q, k, v, s, g = map(lambda x: x.view(-1, C, x.shape[-1]), (q, k, v, s, g))
    # local compute of cumulative product of decay
    # [NC, C, K]
    g = g.cumsum(1)
    ok = gsa_fwd_k(q, k, s, g, M)
    qv = ok.softmax(-1)
    o = gsa_fwd_v(qv, s, v, g, M)
    return o.view(T, -1)
\end{minted}
    \caption{
        Pseudo PyTorch-style code snippet for GSA with chunkwise parallelism.
        For brevity, we omit the dimensions of batch size and number of heads.
        Notably, unlike Algorithm~\ref{alg:gsa}, we obtain the intra outputs via a secondary chunking strategy in Line 31-52 and Line 75-94, as utilized by GLA \cite{yang-etal-2024-gla}, to ensure numerical stability.
    }
    \label{listing:code}
\end{longlisting}

\section{Experimental Setup}\label{sec:setup}
\subsection{Language Modeling}
We compare GSA with the following strong Transformers with modern architectural recipes as well as other recent subquadratic architectures:
\begin{itemize}[leftmargin=15pt]
    \item Xfmr++~\citep{touvron-2023-llama}: Llama-like architectures that enhance the vanilla Transformer by using Rotary position embeddings~\citep{su-2023-roformer} and GLU~\citep{shazeer-2020-glu}.
    \item Mamba~\citep{gu-2023-mamba}: State-space models with \emph{data-dependent} decay.
    \item RetNet~\citep{sun-2023-retnet}: Linear attention with non-learnable, \emph{data-independent} \emph{head-wise} decay and rotary embedding.
    \item GLA~\citep{yang-etal-2024-gla}: Linear attention with elementwise \emph{data-dependent} decay.
    \item HGRN2~\citep{qin-2024-hgrn2}: Gated Linear RNN with state expansion, or GLA with improved parameterization.
\end{itemize}

\paragraph{Setup.}
For a fair comparison, all models are trained from scratch with the same training recipes.
We utilize a subset of 100B tokens picked from the Slimpajama dataset~\citep{cerebras-2023-slimpajama}. The input tokens are processed using the Mistral tokenizer~\citep{jiang-2023-mistral}~\footnote{\url{https://huggingface.co/mistralai/Mistral-7B-v0.1}}.
We use AdamW~\citep{loshchilov-2019-decoupled} with a weight decay $0.01$ as the optimizer.
During training, the learning rate is first warmed up to $3 \times 10^{-4}$ in the first 1B tokens, and then decayed to $3 \times 10^{-5}$ gradually with a cosine schedule.
The number of attention heads is set to 4 and 5 for 1.3B and 2.7B models, respectively.
The number of memory slots is uniformly set to 64 for all models.
We utilize the open-sourced Triton-based library \textsc{FLA} \citep{yang-2024-fla} to run all compared models.

We ran all models on 32 Nvidia H800 GPUs.
To facilitate distributed training and accelerate the process, we utilized the DeepSpeed framework and fused all necessary modules, including ROPE, cross-entropy, and LayerNorm, following the practice of \cite{zhang-2024-tinyllama}. The training of a \textsc{GSA} model with 2.7B parameters took approximately 2 days, while the 1.3B model required 1 day to complete training.

\paragraph{Remark on state size.}

Let the model dimension be denoted as \(d\). Mamba expands the value projection to \(2d\) and uses a state expansion ratio of 16, resulting in a state size of \(32d\) per layer. Since Mamba also replaces the FFN with a Mamba layer, this effectively doubles both the number of recurrent layers and the state size, leading to a total recurrent state size of \(64Ld\).

Similarly, RetNet expands the value projection to \(2d\) and sets the head dimension of queries/keys to be half that of the value head dimension. RetNet also reduces the number of heads to increase the head dimensions of queries and keys. We fix the query/key head dimension to 256 and adjust the number of heads accordingly, resulting in a recurrent state size of \(512d\) per layer and \(512Ld\) in total.

GLA does not expand the value projection but reduces the head dimensions of queries and keys to half of the value head dimension to save parameters for the $\operatorname{Swish}$ output gate, ensuring each layer contains \(4d^2\) parameters. We fix the query/key head dimension to 256 and adjust the number of heads accordingly, resulting in a recurrent state size of \(256d\) per layer and \(256Ld\) in total.

HGRN2 follows a similar approach to GLA but without the $\operatorname{Swish}$ output gate, keeping the head dimensions of queries/keys and values equal, as in standard $\operatorname{softmax}$ attention, while still retaining \(4d^2\) total parameters per recurrent layer. We set the head dimension to 128, resulting in a recurrent state size of \(128d\) per layer and \(128Ld\) in total.

\textsc{GSA} maintains hidden states for both keys and values, so each layer contains a recurrent state size of \(2 \times 64 \times d\). We fix the state expansion (i.e., number of slots) to 64\footnote{Note that in this case, the number of heads is independent of the state expansion ratio}, resulting in a total recurrent state size of \(128Ld\).

\end{document}